\documentclass{article}

 \usepackage[preprint]{neurips_2026}

\usepackage[utf8]{inputenc} % allow utf-8 input
\usepackage[T1]{fontenc}    % use 8-bit T1 fonts
\usepackage{hyperref}       % hyperlinks
\usepackage{url}            % simple URL typesetting
\usepackage{booktabs}       % professional-quality tables
\usepackage{amsfonts}       % blackboard math symbols
\usepackage{nicefrac}       % compact symbols for 1/2, etc.
\usepackage{microtype}      % microtypography
\usepackage{xcolor}         % colors

\usepackage{xspace}
\usepackage{multirow}   % \multirow
\usepackage{graphicx}   % \resizebox, \scalebox
\usepackage{amsmath}

\usepackage{subcaption}
\usepackage{makecell}
\usepackage{tabularx}

\usepackage[table]{xcolor}
\definecolor{baselinecolor}{gray}{.9}
\newcommand{\gray}[1]{\cellcolor{baselinecolor}{#1}}

\newcommand{\model}{\textbf{ConCor-1}\xspace}

\title{Vision-Language Grounding as Bidirectional Concept Correspondence}

\author{%
  Jieyu Zhang\textsuperscript{1}\thanks{Equal contribution.}
  \quad
  Ziqi Gao\textsuperscript{1,2}\footnotemark[1]
  \quad
  Luke Zettlemoyer\textsuperscript{1,3}
  \quad
  Ranjay Krishna\textsuperscript{1}
  \\[4pt]
  \textsuperscript{1}University of Washington
  \\
  \textsuperscript{2}Allen Institute for AI
  \\
  \textsuperscript{3}FAIR at Meta
  \\[20pt]
  \textbf{Project Page:} \href{https://uwgzq.github.io/papers/ConCor-1}{\texttt{https://uwgzq.github.io/papers/ConCor-1}}
}

\begin{document}

\maketitle
\vspace{-1.5em}

\begin{abstract}

Vision-language grounding connects language to visual content, yet most existing formulations reduce grounding to a unidirectional localization problem; given a prespecified text phrase or category name, identify the corresponding image region. This setup assumes that the relevant linguistic unit is already known, overlooking a more basic challenge in grounded communication: determining which parts of the text are visually referential and how they correspond to entities in the image. 
We formulate grounding as \textit{bidirectional concept correspondence} over an image-text pair. Given an image and its paired text, the goal is to recover all correspondences between visually referential text spans and instance-level image segments, without assuming that the relevant text spans are provided. 
This formulation unifies common grounding tasks, including phrase grounding, referring expression grounding, and open-vocabulary detection by treating text segmentation, image segmentation, and cross-modal alignment as a single correspondence prediction problem. 
To address this task, we introduce \model, a grounding model built on top of a pretrained vision-language model. 
It uses learnable \textit{bridge tokens} to represent candidate image-text correspondences and predicts, for each token, a text mask, an image mask, and a correspondence presence score. To train and evaluate this task, we convert diverse grounding and segmentation datasets into a unified correspondence format. 
Experiments show that \model consistently outperforms baselines, improving correspondence F1 by 48\% on the long-caption dataset and by 29\% on zero-shot LVIS, where the large category list serves as the text input.

\end{abstract}

\section{Introduction}

Grounding is a long-standing problem in vision-language learning that asks how linguistic concepts correspond to visual content in an image.
% As a core interface between language and perception, grounding underlies a wide range of applications, including visual content editing, assistive perception, and embodied agents. 
Grounding enables multimodal systems to produce outputs that are spatially attributable, making their predictions more interpretable and actionable~\cite{molmov1,clark2026molmo2openweightsdata,clark2025molmopoint}.

Most existing formulations, however, simplify grounding into a unidirectional language-to-image localization task. 
Early work studied referring expressions~\cite{yu2016modeling,mao2016generation,nagaraja2016modeling,krishna2018referring}, where a short phrase such as ``squirrel'' identifies a target object, as in ReferItGame~\cite{kazemzadeh2014referitgame}. 
Later work expanded grounding to phrase grounding~\cite{plummer2015flickr30k,wu2020phrasecut}, where referential phrases in a caption, such as ``a brown dog'' or ``the wooden bench'' are aligned to image regions, and further to open-vocabulary detection and segmentation~\cite{liu2023grounding,carion2025sam3segmentconcepts}.
These tasks typically assume that the relevant linguistic unit is already specified. 
Text indicates what to find, and the image provides the region to retrieve.

Grounding is not simply a one-way retrieval process. 
In human communication, reference is a joint activity between speaker and listener, where utterances are interpreted against shared perceptual context and common ground~\cite{clark1996using,clark1991grounding}. 
Listeners are not handed a pre-segmented list of phrases to localize; rather, they must infer which parts of an utterance are referential, which entities in the scene are relevant, and how the two should be aligned. 
Likewise, cognitive accounts of spatial language show that people construct structured spatial mental models from descriptions, rather than treating words as isolated category labels~\cite{tversky1991spatial,taylor1992spatial}. 
Visual cognition also requires selective attention~\cite{eftekhar2023selective} and object binding: image regions must be grouped into a coherent structure before they can serve as referents for language~\cite{treisman1980feature}. 
Together, these perspectives suggest that visual grounding is more naturally viewed as a problem of recovering correspondences between structured linguistic and visual representations~\cite{gentner1983structure}.

Motivated by this perspective, we formulate vision-language grounding as \textit{bidirectional concept correspondence} over an image-text pair. 
In other words, grounding determines both \textit{what in the text is visually referential} and \textit{what in the image it refers to}. 
% For example, in a caption containing multiple entities, modifiers, repeated mentions, and pronouns, the model must decide which spans refer to visible entities, group coreferential mentions, and align each group to the corresponding visual instance. 
Given an image and its paired text, the goal is to recover the complete set of correspondence pairs $\{(t_i, s_i)\}$, where $s_i$ is an instance-level image mask and $t_i$ is a binary mask over text marking all mentions that refer to the same visual entity. 
% Unlike conventional phrase grounding, the relevant text spans are not provided in advance. 
The model must jointly identify grounded text segments, segment the corresponding visual entities, and determine whether each proposed text-image pairing is valid. 
Moreover, $t_i$ may contain multiple non-contiguous spans, allowing the model to group repeated or coreferential mentions.

This formulation subsumes several existing grounding tasks as special cases. 
Phrase grounding arises when the relevant text spans are already specified in the paired caption. 
Open-vocabulary detection and segmentation correspond to the restricted case where the text consists only of category names. 
Referring expression grounding corresponds to the case where a query expression is provided, but our formulation additionally requires the model to ground contextual entities mentioned in the expression. 
More broadly, bidirectional concept correspondence connects vision-language grounding to visual coreference resolution~\cite{goel2023you} and Winograd-style referential ambiguity~\cite{winograd1972understanding,levesque2012winograd,park2024picturing}, where resolving expressions such as ``it'' in ``The trophy does not fit into the suitcase because it is too large'' depends on linking language to the visual or physical situation.

\begin{figure}[t!]
    \centering
\includegraphics[width=\textwidth]{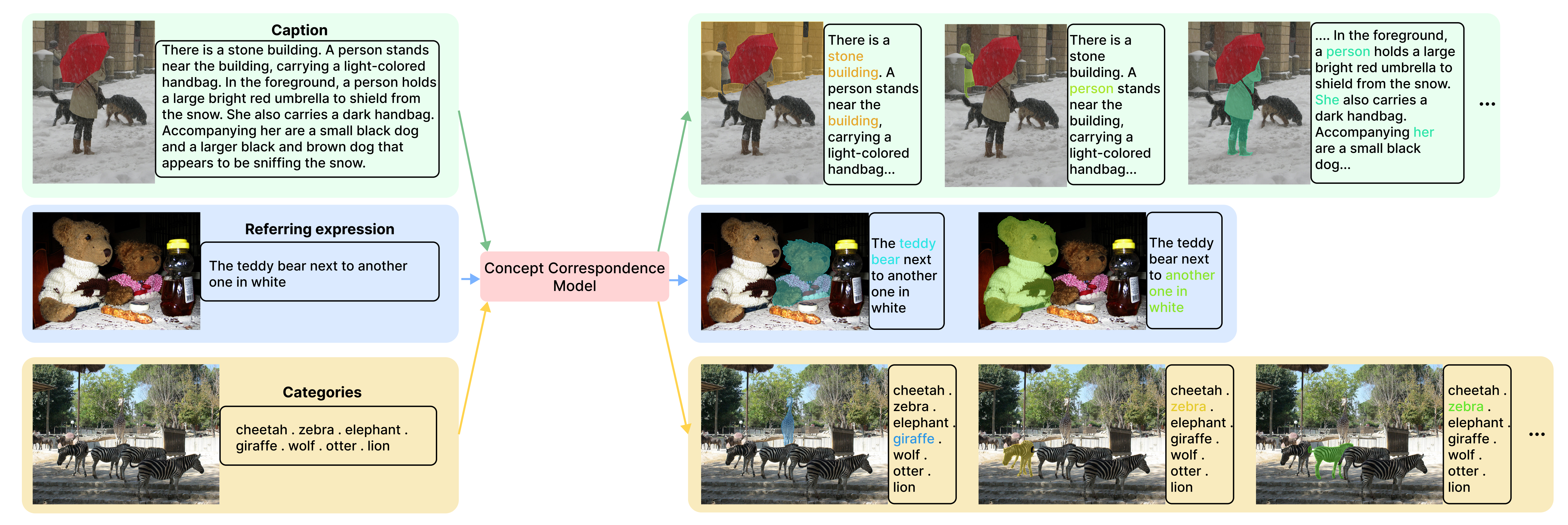}
    \caption{
\textbf{Bidirectional concept correspondence.}
Given an image paired with different forms of text, including a full caption, a referring expression, or a list of category names, the concept correspondence model predicts a complete set of correspondences between text segments and image segments.
For each correspondence, the model identifies the visually grounded text span, including repeated or coreferential mentions, and localizes the corresponding instance-level mask in the image.
% This unified formulation subsumes caption grounding, referring expression grounding, and open-vocabulary segmentation under the same output space of text--image correspondence pairs.
}
\vspace{-6mm}
    \label{fig:task}
\end{figure}

% \ranjay{Move this to related work section:}
% Although prior work has explored related ideas—from early region-text alignment models~\cite{karpathy2015deep} to recent grounding-based systems such as MDETR~\cite{kamath2021mdetr}, DetClip~\cite{yao2022detclip, yao2023detclipv2}, GLIP~\cite{li2022grounded,zhang2022glipv2}, and Grounding DINO~\cite{liu2023grounding}—these approaches have largely treated correspondence as a means rather than an end. In most cases, the goal is not to recover the full set of grounded text-image pairs, but to improve downstream tasks such as detection or caption generation. In contrast, we treat bidirectional concept correspondence as the task itself. 

To address this task, we introduce \model, a correspondence prediction model built on top of a pretrained vision-language model (we use Qwen3.5~\cite{qwen35} as our backbone). 
This design is motivated by two requirements of bidirectional correspondence. 
First, recovering all grounded entities in an image-text pair requires reasoning over long and compositional text, including modifiers, relations, repeated mentions, and discourse context. 
Second, because text segmentation is part of the output space, the model must identify visually grounded spans, resolve coreference, and interpret contextual references. 
Large pretrained vision-language models provide a natural backbone for these capabilities because they jointly process image and text while preserving a flexible language interface.

Built on this shared backbone, \model adds three lightweight prediction heads: one for image segmentation, one for text segmentation, and one for correspondence presence prediction. 
To represent individual correspondences, we introduce a set of learnable \textit{bridge tokens} that act as queries for potential correspondences. 
We append these bridge tokens to the multimodal token sequence, allowing them to be contextualized through self-attention over both image tokens and text tokens in the pretrained backbone.
After interacting with both vision and text tokens, the image segment and text mask are predicted by measuring similarity between the bridge token and the vision or text tokens, respectively, while the presence head predicts whether the correspondence is valid.
This design keeps the pretrained backbone largely intact while adding a minimal task-specific interface for predicting bidirectional correspondences.

We train and evaluate \model by converting diverse grounding, detection, and segmentation datasets into a unified correspondence format.
Experiments show that \model consistently performs best at predicting concept correspondences across two evaluation settings: image-caption grounding, where the model must identify visually referential spans in natural captions, and image-category grounding, where the input text is a large list of candidate categories.
The gains are especially substantial on COCONut-PanCap, which contains long, complex captions with dense object references: compared with the strongest baseline, \model improves correspondence F1 (JointF1) by 48\% and correspondence IoU (mJS) by 41\%.
In the zero-shot, large-vocabulary LVIS setting, \model is able to process the full category vocabulary in a single forward pass and outperforms the strongest baseline by 29\% in JointF1.
These results show that explicit bridge-token correspondence prediction enables a unified model to recover text spans, image masks, and their alignments across both dense captions and large-vocabulary category queries.

\section{Related work}

\textbf{Vision-language grounding.}
Vision-language grounding has been studied under several formulations that differ mainly in how the linguistic query is specified and what visual output is required~\cite{clark2025molmopoint,yao2026qwen3,zheng2025one,zheng2026trajtok,wei2026youtu}. Early work focused on referring expression comprehension in natural images, where a short phrase is used to localize a target object or region~\cite{kazemzadeh2014referitgame,yu2016modeling,mao2016generation,nagaraja2016modeling}. Later, phrase grounding extended this setup to align multiple phrases in a caption with image regions~\cite{plummer2015flickr30k,wu2020phrasecut}, while more recent open-vocabulary detection and segmentation methods allow arbitrary category names or free-form text queries to localize boxes or masks~\cite{li2022grounded,zhang2022glipv2,liu2023grounding,park2025synthetic,gao2026syntheticvisualgenome2}. Despite these differences, most prior formulations remain fundamentally \emph{query-driven}: text is treated as an input specification for retrieval, and the objective is to localize the corresponding visual region. In contrast, we study grounding as a \emph{bidirectional concept correspondence} problem over an image-text pair, where the model must recover the full set of aligned text-image concepts without assuming that the grounded text spans are given in advance.

\textbf{Grounding models.}
Prior work on grounding and segmentation includes detector-style methods such as MDETR~\cite{kamath2021mdetr}, DetCLIP~\cite{yao2022detclip,yao2023detclipv2}, GLIP~\cite{li2022grounded,zhang2022glipv2}, Grounding DINO~\cite{liu2023grounding,zhao2024open,fu2025llmdet}, and GLEE~\cite{wu2024general}; generalized and promptable segmentation models such as X-Decoder~\cite{zou2023generalized}, OpenSeeD~\cite{zhang2023simple}, SEEM~\cite{zou2023segment}, SAM3~\cite{carion2025sam3segmentconcepts}, and OpenWorldSAM~\cite{xiao2025openworldsam}; and autoregressive VLM-based systems including Florence-2~\cite{xiao2024florence}, Kosmos-2~\cite{peng2023kosmos2}, Molmo~\cite{molmov1,clark2026molmo2openweightsdata}, LISA~\cite{lai2023lisa,yang2023lisa++}, and GLaMM~\cite{rasheed2024glamm}.
In contrast, our model combines extensively pretrained VLM backbones and segmentation-based grounding: it builds on a pretrained VLM for rich joint image-text understanding, while adding lightweight segmentation-oriented heads to explicitly predict text spans, image segments, and their correspondences in a non-autoregressive and structured manner.

% \subsection{Problem formulation}

% Given an image and its paired text, the goal is to recover the complete set of concrete correspondences between \emph{text segments} and \emph{image segments} that refer to the same visual entity. 
% Unlike conventional grounding settings, the relevant text spans are not provided as input queries; instead, the model must identify which parts of the text are visually grounded and align them with the corresponding image regions.

\section{Bidirectional concept correspondence}

We start by defining vision-language grounding as a bidirectional correspondence problem between language and vision. 
Formally, given an image $I$ and a text sequence $T$, the output is a set of correspondence pairs $\mathcal{C}=\{(t_i, s_i)\}_{i=1}^{K}$,
where each $t_i$ denotes a grounded text segment in $T$ and each $s_i$ denotes the corresponding image segment in $I$. 
Each $t_i$ is not necessarily restricted to a single contiguous character span; it can be represented as a binary mask over characters in the text, allowing the formulation to capture discontinuous or multi-span grounded expressions.

By requiring the model to recover the complete set of concrete correspondences in an image–text pair, the model must first interpret the text to resolve coreference and identify entity mentions, then examine the image to surface candidate objects, and finally determine which mentions and objects correspond—while recognizing that some mentions may have no visual referent and some objects may not be described in the text. This richer structure also enables applications beyond standard grounding benchmarks, including evaluating the groundedness of image captions and the faithfulness of text-to-image generations.

This new task poses key technical challenges. The model must handle long, compositional text with complex referring structures, including modifiers, repeated mentions, and pronouns. It must also produce fine-grained instance image masks with local spatial detail. Finally, a single text–pixel similarity matrix is insufficient because it only provides local affinities, not a structured set of correspondences. The model must decide how to group tokens into referential text masks, how to group pixels into instance-level image masks, how many correspondences exist, and which text and image groups should be paired.

% \ranjay{Here is where you want to lay out the technical challenges: mention with an example that text spans do not have to be continuous. That image segments need to be predicted with high granularily, which is often collapsed or lost when image patches are tokenized.}

\subsection{Concept Correspondence Model}

\begin{figure}[t!]
    \centering
\includegraphics[width=\textwidth]{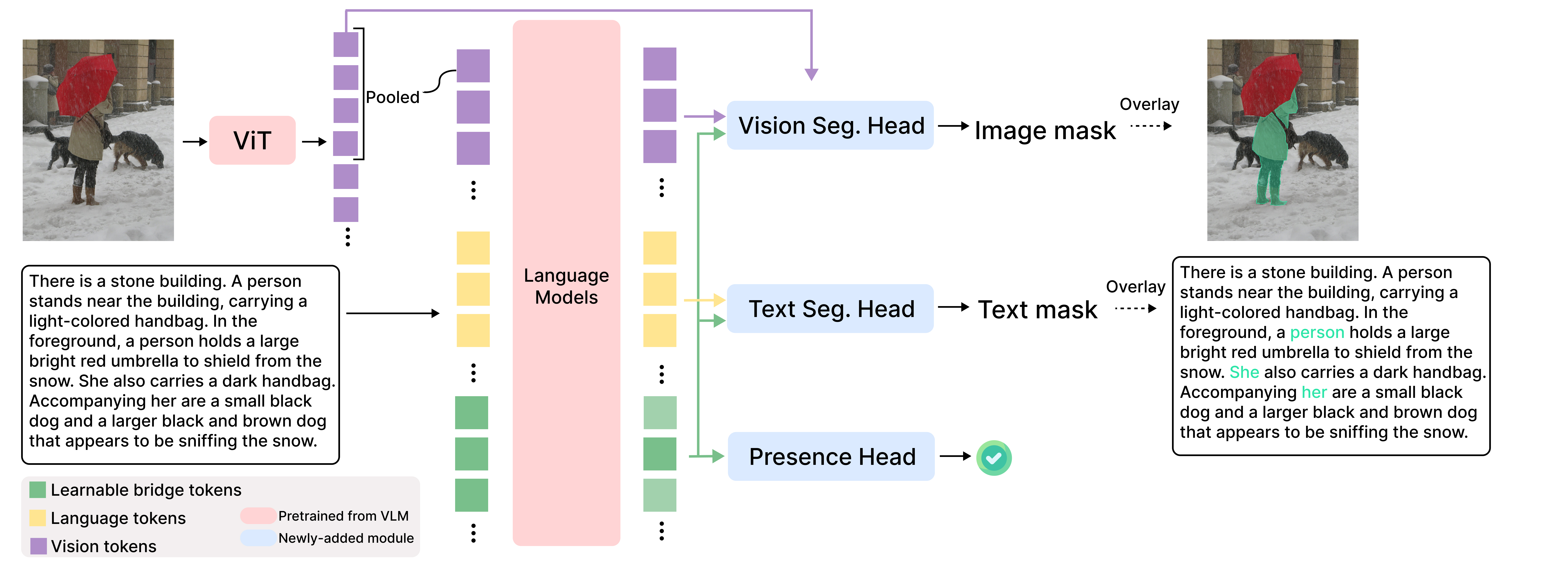}
    \caption{
\textbf{Model overview.}
\model builds on a pretrained vision-language backbone and introduces learnable bridge tokens to represent candidate text--image correspondences.
Given an image and paired text, vision tokens, language tokens, and bridge tokens are jointly processed by the language model.
For each bridge token, the model predicts an image mask with a vision segmentation head, a text mask with a text segmentation head, and a presence score indicating whether the token corresponds to a valid grounded concept.
}
\vspace{-3mm}
    \label{fig:model}
\end{figure}

We build the model (\model) on top of a pretrained vision-language backbone (Qwen3.5~\cite{qwen35}). 
This choice is motivated by its strong long-context modeling, rich text understanding (e.g., coreference and compositionality), and broad multimodal knowledge acquired during pretraining, all of which are essential for identifying and aligning vision--text concepts without predefined text spans.
We use the 0.8B variant to keep the model size comparable to existing grounding models.
Although Qwen3.5 is originally designed for autoregressive language generation, we instead use it as a contextual image--text encoder and attach lightweight prediction heads to directly predict text masks, image masks, and correspondence presence.

As shown in Fig.~\ref{fig:model}, given an image--text pair, the backbone first produces a unified sequence of visual and textual tokens that encode both modalities in a shared representation space. 
We additionally introduce a set of learned \emph{bridge tokens}, where each bridge token represents a candidate image--text correspondence. 
By attending jointly to visual and textual tokens, each bridge token aggregates the context needed to hypothesize a potential alignment between an image region and a text span. 
Conditioned on this representation, the model simultaneously predicts (1) an image mask for the corresponding visual segment, (2) a text mask for the corresponding text segment, and (3) a presence score indicating whether the proposed pairing is a valid correspondence.

\textbf{Bridge token assignment.}
Because bridge tokens are fixed in number and initially unordered, while each image--text pair contains a variable number of correspondences, the model needs a training-time assignment strategy that encourages different tokens to specialize to different correspondence hypotheses.
A simple approach is to follow raster order over image objects~\cite{molmov1}, but this provides no explicit structural prior.
Instead, we introduce a multi-scale spatial assignment strategy: we allocate bridge tokens over grids at multiple resolutions, with token groups corresponding to partitions such as \(1\), \(2\times2\), \(3\times3\), and so on over the image.
Each bridge token is therefore associated with a spatial cell at a particular scale.
During training, we compute the IoU between each ground-truth image mask and each candidate grid cell, and then use Hungarian matching to assign correspondences to bridge tokens.
This gives bridge tokens both spatial and scale priors, encouraging them to specialize in objects with different locations and extents.

\textbf{Presence head.}
For each bridge token, we predict a scalar \emph{presence score} indicating whether it corresponds to a valid image--text concept pair. This head is implemented as a lightweight MLP applied to the final hidden state of the bridge token. 

\textbf{Text segmentation head.}
To recover the grounded text segment associated with a bridge token, we predict a binary mask over the text tokens in the input sequence. 
% Concretely, let $\mathbf{b}_j$ denote the representation of the $j$-th bridge token and let $\mathbf{x}_\ell^{\text{text}}$ denote the representation of the $\ell$-th text token from the backbone. 
We first project bridge and text token features into a shared correspondence space, and then score each bridge--text-token pair with a learnable bilinear function, producing a logit for whether the token belongs to the grounded text segment represented by that bridge token. 
% Applying this independently across all text tokens yields a binary text mask for each bridge token. This formulation naturally supports multi-token and discontinuous text spans, since the model predicts token-level membership rather than forcing a contiguous span boundary.

\textbf{Image segmentation head.}
% To recover the corresponding image segment, we similarly predict a binary mask over image patches for each bridge token. Because the backbone visual tokens are spatially compressed through $2\times2$ patch merging, we first map them back to a denser patch-level representation using learned projection layers, and fuse them with patch-level features from the vision encoder to restore fine-grained local detail. The fused features are then processed by a lightweight convolutional upsampling decoder to produce a high-resolution feature map. Similar to the text segmentation head, we use a learnable bilinear scoring function that takes as input the projected bridge token feature and the high-resolution feature map to predict a binary mask over spatial locations.
To recover the corresponding image mask, we predict a binary pixel mask for each bridge token. Because the vision tokens in language models are spatially compressed through $2\times2$ patch merging, we first project them back to the original patch grid using learned projection layers. We then fuse these reconstructed VLM features with patch-level features from the vision encoder to restore local spatial detail. The fused features are processed by a lightweight convolutional decoder to produce a spatial feature map at the mask-prediction resolution. Similar to the text segmentation head, we project the bridge token and spatial features into a shared correspondence space, and use a learnable bilinear scoring function to predict whether each spatial location belongs to the image mask associated with that bridge token.

\textbf{Training objectives.}
We train the model to jointly predict text masks, visual masks, and correspondence validity for each bridge token. Text masks are supervised with binary cross-entropy (BCE). For visual masks, we use a PointRend-style~\cite{kirillov2020pointrend} point-sampled loss, combining BCE and Dice loss~\cite{milletari2016vnet} over matched positive bridges. A separate BCE loss supervises the presence head, indicating whether each bridge token corresponds to a valid text--image pair.
A more detailed description of the model and training implementation can be found in the appendix.

\subsection{Concept correspondence dataset and evaluation}

We convert diverse grounding, segmentation, and referring-expression datasets into a unified correspondence format with image masks, text masks, and correspondence labels, with additional dataset details and statistics provided in the appendix.

\textbf{Repurposing existing grounded-caption datasets.}
Existing grounded-caption datasets are valuable but not directly compatible with our bidirectional correspondence formulation. Their annotations are often incomplete: co-referring text mentions are typically annotated only at first occurrence, and some mentioned entities lack corresponding image masks~\cite{liu2023grounding,deng2025coconut,plummer2015flickr30k}. We therefore use an LLM-based rewriting pipeline that groups co-referring mentions into shared text masks and removes mentions without associated image masks. This yields complete correspondence annotations between coreference-aware text masks and instance-level image masks.

\textbf{Training data.}
We include the following repurposed or constructed datasets as training data.
GoldG~\cite{liu2023grounding} and COCONut-PanCap~\cite{deng2025coconut} are built from human-written captions and processed with the aforementioned LLM rewriting pipeline.
GoldG is derived from Flickr30k~\cite{plummer2015flickr30k} and GQA~\cite{hudson2019gqa} annotations from the Grounding DINO training corpus, with pseudo masks generated from ground-truth boxes using SAM3.
COCONut-PanCap uses relabeled COCO panoptic captions with ground-truth masks.
We also construct GroundedRef-train from COCO panoptic segmentation by generating referring and compositional captions with an LLM and pairing them with ground-truth masks.
Along with these caption-style datasets, we incorporate instance segmentation datasets, including COCO~\cite{lin2015microsoftcococommonobjects}, COCONut~\cite{coconut2024cvpr}, EntitySeg~\cite{qi2023highquality},
% LVIS~\cite{gupta2019lvis}, 
and ADE20K~\cite{zhou2019semantic}.
We further include the counting subset of PixMo-Point~\cite{molmov1}, converting points to masks with SAM3~\cite{carion2025sam3segmentconcepts}.
Overall, our training set contains 653K image--text pairs over 184K unique images for caption-style datasets, annotating 3.1M correspondences over 4.2M instance masks, and 614K images with 7.8M instance masks for instance segmentation data.

\textbf{Evaluation data.}
For evaluation on image--caption pairs, we process the Flickr30k~\cite{plummer2015flickr30k} and COCONut-PanCap~\cite{deng2025coconut} validation sets using the same LLM rewriting pipeline, followed by human filtering to ensure annotation quality.
We also construct a separate GroundedRef validation set from the validation splits of RefCOCOg~\cite{mao2016generation} and gRefCOCO~\cite{he2023grec}.
Because these datasets primarily annotate only the target object referred to by each expression, we additionally annotate masks for contextual objects mentioned in the expression, using SAM3 proposals followed by manual verification.
This makes the validation examples compatible with our task.
The final evaluation sets contain 2K image--caption pairs for Flickr30k, 2.2K for COCONut-PanCap, and 2K for GroundedRef.
% \clearpage
\section{Experiments}

We train the \model on 8 H100 GPUs for 100K optimization steps with an effective batch size of 96, using image cropping and collage data augmentations. Further implementation details are provided in the appendix.
Our experiments evaluate bidirectional concept correspondence across image--caption and image--category settings. 
% Overall, \model achieves the strongest joint correspondence results on caption-based benchmarks, remains competitive on COCO, and shows clear gains on LVIS, where it can process the full category vocabulary in a single forward pass.

\subsection{Main results}

\begin{table*}[t]
\centering
\scriptsize
\caption{Image-caption results on COCONut-PanCap, GroundedRef, and Flickr30k validation sets.}
\label{tab:grounding-main-results}
\setlength{\tabcolsep}{2pt}
\renewcommand{\arraystretch}{0.9}
\resizebox{\textwidth}{!}{
\begin{tabular}{lcccccc|cccccc|cccccc}
\toprule

& \multicolumn{6}{c|}{\scriptsize\textbf{COCONut-PanCap}}
& \multicolumn{6}{c|}{\scriptsize\textbf{GroundedRef}}
& \multicolumn{6}{c}{\scriptsize\textbf{Flickr30k}} \\

\cmidrule(lr){2-7}
\cmidrule(lr){8-13}
\cmidrule(lr){14-19}

{\scriptsize\textbf{Method}}
& \rotatebox{45}{\tiny \textbf{JointF1}}
& \rotatebox{45}{\tiny \textbf{MaskF1}}
& \rotatebox{45}{\tiny \textbf{TextF1}}
& \rotatebox{45}{\tiny \textbf{mJS}}
& \rotatebox{45}{\tiny \textbf{mMaskIoU}}
& \rotatebox{45}{\tiny \textbf{mSpanIoU}}

& \rotatebox{45}{\tiny \textbf{JointF1}}
& \rotatebox{45}{\tiny \textbf{MaskF1}}
& \rotatebox{45}{\tiny \textbf{TextF1}}
& \rotatebox{45}{\tiny \textbf{mJS}}
& \rotatebox{45}{\tiny \textbf{mMaskIoU}}
& \rotatebox{45}{\tiny \textbf{mSpanIoU}}

& \rotatebox{45}{\tiny \textbf{JointF1}}
& \rotatebox{45}{\tiny \textbf{MaskF1}}
& \rotatebox{45}{\tiny \textbf{TextF1}}
& \rotatebox{45}{\tiny \textbf{mJS}}
& \rotatebox{45}{\tiny \textbf{mMaskIoU}}
& \rotatebox{45}{\tiny \textbf{mSpanIoU}}
\\

\midrule

GDINO+SAM
& 34.1 & 64.8 & 39.8 & 48.0 & 67.9 & 44.2 
& 45.0 & 56.4 & 50.9 & 37.8 & 44.2 & 39.7 
& 81.6 & 85.2 & 86.7 & 81.9 & 86.9 & 81.5 \\

MM-GDINO+SAM
& 32.2 & 59.5 & 37.4 & 51.1 & 70.8 & 47.4 
& 39.8 & 58.1 & 44.3 & 31.9 & 44.2 & 33.3 
& 84.6 & 87.7 & 89.6 & 82.6 & 87.3 & 82.3 \\ 

LLMDet+SAM
& 31.5 & 52.8 & 37.9 & 59.1 & 75.4 & 55.7 
& 54.6 & 69.0 & 62.8 & 68.1 & \textbf{77.9} & 70.1 
& 76.7 & 79.6 & 85.4 & 75.3 & 78.7 & 77.1 \\ 

GPT-5.4 (medium)+SAM
& 43.8 & 47.4 & 88.9 & 48.6 & 43.5 & 82.7
& 25.5 & 37.5 & 68.2 & 33.9 & 34.3 & 70.8
& 37.3 & 43.4 & 86.6 & 44.7 & 41.4 & 87.0 \\

Florence-2
& 36.7 & 53.4 & 53.6 & 47.9 & 55.8 & 56.7 
& 52.6 & 61.2 & 79.2 & 56.4 & 55.2 & 77.7 
& 68.7 & 70.1 & 92.8 & 69.4 & 61.6 & 88.9 \\

GLaMM
& 2.5 & 39.1 & 3.0 & 11.9 & 27.4 & 8.0
& 6.1 & 47.3 & 10.0 & 18.4 & 32.8 & 16.9
& 53.1 & 69.1 & 59.5 & 55.4 & 60.3 & 58.2 \\

Qwen3.5-FT
& 59.9 & 63.4 & 93.1 & 63.3 & 55.6 & 87.9
& 51.5 & 56.3 & \textbf{82.1} & 49.7 & 45.8 & 77.5
& 70.2 & 72.6 & \textbf{95.2} & 69.1 & 60.3 & 95.1 \\

\midrule
\model{} (random init.)
& 51.8 & 59.4 & 67.6 & 63.2 & 63.6 & 77.9
& 32.4 & 43.4 & 56.9 & 44.6 & 48.1 & 70.5
& 55.0 & 61.1 & 72.7 & 64.4 & 62.3 & 83.7 \\

\model
& \gray{\textbf{88.8}} & \gray{\textbf{91.9}} & \gray{\textbf{93.4}} & \gray{\textbf{89.5}} & \gray{\textbf{87.0}} & \gray{\textbf{95.7}}
& \gray{\textbf{70.3}} & \gray{\textbf{76.4}} & \gray{78.7} & \gray{\textbf{69.6}} & \gray{68.8} & \gray{\textbf{78.7}}
& \gray{\textbf{91.4}} & \gray{\textbf{92.4}} & \gray{95.1} & \gray{\textbf{91.8}} & \gray{\textbf{88.6}} & \gray{\textbf{97.4}} \\

\bottomrule
\end{tabular}
}
\vspace{-10pt}
\end{table*}

\begin{table*}[t]
\centering
\scriptsize
\caption{Image-category results on COCO validation set, LVIS-minival, and EntitySeg validation set.}
\label{tab:instance-seg-results}
\setlength{\tabcolsep}{2pt}
\renewcommand{\arraystretch}{0.9}
\resizebox{\textwidth}{!}{
\begin{tabular}{lcccccc|cccccc|cccccc}
\toprule

% & \multicolumn{6}{c|}{\textbf{COCO}}
% & \multicolumn{6}{c|}{\textbf{LVIS-minival}}
% & \multicolumn{6}{c}{\textbf{EntitySeg}} \\
& \multicolumn{6}{c|}{\scriptsize\textbf{COCO}}
& \multicolumn{6}{c|}{\scriptsize\textbf{LVIS-minival}}
& \multicolumn{6}{c}{\scriptsize\textbf{EntitySeg}} \\

\cmidrule(lr){2-7}
\cmidrule(lr){8-13}
\cmidrule(lr){14-19}

{\scriptsize\textbf{Method}}
& \rotatebox{45}{\tiny \textbf{JointF1}}
& \rotatebox{45}{\tiny \textbf{MaskF1}}
& \rotatebox{45}{\tiny \textbf{TextF1}}
& \rotatebox{45}{\tiny \textbf{mJS}}
& \rotatebox{45}{\tiny \textbf{mMaskIoU}}
& \rotatebox{45}{\tiny \textbf{mSpanIoU}}

& \rotatebox{45}{\tiny \textbf{JointF1}}
& \rotatebox{45}{\tiny \textbf{MaskF1}}
& \rotatebox{45}{\tiny \textbf{TextF1}}
& \rotatebox{45}{\tiny \textbf{mJS}}
& \rotatebox{45}{\tiny \textbf{mMaskIoU}}
& \rotatebox{45}{\tiny \textbf{mSpanIoU}}

& \rotatebox{45}{\tiny \textbf{JointF1}}
& \rotatebox{45}{\tiny \textbf{MaskF1}}
& \rotatebox{45}{\tiny \textbf{TextF1}}
& \rotatebox{45}{\tiny \textbf{mJS}}
& \rotatebox{45}{\tiny \textbf{mMaskIoU}}
& \rotatebox{45}{\tiny \textbf{mSpanIoU}}
\\

\midrule

GDINO+SAM
& 63.0 & 66.5 & 73.3 & 76.0 & 71.3 & 90.7
& 15.3 & 22.4 & 17.7 & 33.1 & 47.8 & 35.7
& 21.5 & 30.6 & 27.0 & 36.9 & 45.5 & 43.3 \\

MM-GDINO+SAM
& 65.7 & 68.7 & 75.6 & 77.1 & 72.0 & 91.6
& 9.4 & 18.4 & 12.4 & \textbf{53.6} & 75.7 & \textbf{56.8}
& 29.4 & 36.9 & 37.0 & 45.5 & 51.7 & 55.7 \\

LLMDet+SAM
& 62.4 & 65.9 & 72.1 & 77.2 & 72.2 & 91.9
& 4.3 & 8.7 & 5.8 & 53.2 & \textbf{76.0} & 56.5
& 22.2 & 31.7 & 29.2 & 44.5 & \textbf{56.2} & 55.2 \\

GPT-5.4 (medium)+SAM
& 34.2 & 35.1 & \textbf{81.7} & 39.7 & 32.5 & 80.0
& 17.2 & 20.9 & \textbf{43.0} & 25.8 & 26.3 & 53.7
& 20.2 & 30.1 & 42.4 & 30.7 & 34.4 & 51.9 \\

Florence-2
& 9.8 & 14.8 & 11.9 & 7.6 & 10.1 & 9.1
& 6.2 & 17.7 & 8.2 & 17.0 & 38.0 & 19.6
& 11.8 & 20.9 & 23.5 & 21.2 & 28.8 & 32.1 \\

GLaMM
& 38.5 & 40.8 & 45.7 & 33.1 & 31.5 & 38.9
& 23.1 & 24.0 & 29.7 & 19.4 & 18.4 & 24.1
& 22.2 & 23.1 & 26.0 & 18.3 & 17.4 & 21.2 \\

Qwen3.5-FT
& 39.4 & 41.2 & 51.4 & 37.3 & 37.4 & 48.3
& 3.9 & 26.9 & 5.2 & 4.4 & 28.0 & 5.2
& 31.1 & 38.6 & 47.6 & 33.1 & 38.0 & 43.6 \\

\midrule
\model{} (random init.)
& 15.1 & 32.8 & 22.9 & 25.1 & 49.5 & 35.6
& 0.1 & 18.5 & 0.2 & 0.2 & 32.5 & 0.3
& 2.8 & 20.4 & 5.4 & 4.3 & 26.6 & 6.5 \\

\model
& \gray{\textbf{72.4}} & \gray{\textbf{74.3}} & \gray{81.0}
& \gray{\textbf{79.8}} & \gray{\textbf{74.2}} & \gray{\textbf{92.7}}
& \gray{\textbf{29.9}} & \gray{\textbf{41.4}} & \gray{33.8}
& \gray{39.8} & \gray{51.4} & \gray{45.0}
& \gray{\textbf{49.8}} & \gray{\textbf{59.9}} & \gray{\textbf{59.1}}
& \gray{\textbf{48.9}} & \gray{53.7} & \gray{\textbf{57.4}} \\

\bottomrule
\end{tabular}
}
\vspace{-15pt}
\end{table*}

\textbf{Evaluation metrics.}
We evaluate predictions at three levels: text segmentation, image segmentation, and joint correspondence. For text segmentation, we match predictions to ground truth with Hungarian matching on character span IoU. A match is considered correct if span IoU $\geq 0.5$, and we report \textbf{TextF1}. To capture quality beyond thresholded correctness, we also report \textbf{mSpanIoU}, the average span IoU of matched pairs over all ground-truth groups, with unmatched groups contributing zero. For image segmentation, we apply the same procedure using mask IoU, and report \textbf{MaskF1} and \textbf{mMaskIoU}. For joint correspondence, we perform Hungarian matching on the geometric mean of text and mask IoU, and count a match as correct only when both exceed $0.5$. We report \textbf{JointF1} together with \textbf{mJS}, a soft joint score defined as the average geometric mean over matched pairs, with unmatched ground-truth instances assigned zero.

\textbf{Baselines.}
We compare against open-vocabulary grounding baselines that localize candidate regions and produce grounded mask outputs. GDINO+SAM uses Grounding DINO~\cite{liu2023grounding} to detect caption-conditioned boxes and recover text spans from token similarity scores, then refines boxes into masks with SAM3~\cite{carion2025sam3segmentconcepts}. We also include its successors, MM-GDINO~\cite{zhao2024open} and LLMDet~\cite{fu2025llmdet}. Florence-2~\cite{xiao2024florence} predicts grounded boxes and phrase labels from the caption, recovers character spans by phrase matching, and converts boxes to masks with its region-to-segmentation head. Long captions are chunked when needed for these methods. GLaMM~\cite{rasheed2024glamm} natively generates interleaved grounded phrases and masks; we condition its FullScope checkpoint on the target caption and match generated phrases back to caption spans. GPT-5.4~\cite{openai2026gpt54} (medium reasoning) predicts grounded caption spans or category names with boxes, which SAM3~\cite{carion2025sam3segmentconcepts} converts to masks. Finally, we fine-tune Qwen3.5-0.8B~\cite{qwen35} (Qwen3.5-FT) with autoregressive supervision to output text spans and polygon masks for each correspondence.

\textbf{Results overview.}
Table~\ref{tab:grounding-main-results} reports results on three image-caption benchmarks: COCONut-PanCap, GroundedRef, and Flickr30k.
Table~\ref{tab:instance-seg-results} reports results on three image-category benchmarks: COCO, LVIS-minival, and EntitySeg, where category names are concatenated into a single text input; because the LVIS vocabulary is large, all baselines except Qwen3.5-FT require chunking the category list and running multiple forward passes.
Across both settings, \model consistently achieves the best F1-based metrics, including JointF1, MaskF1, and TextF1, showing that it can recover text spans, image masks, and their alignments across both natural captions and category-list queries.

\textbf{Findings.}
We list several findings from the results.
(1) \emph{Explicit correspondence prediction outperforms language-only generation.}
Qwen3.5-FT formulates the task as autoregressive language generation of text spans and polygon masks, and performs strongly on text metrics, but \model improves JointF1 from 59.9 to 88.8 and mJS from 63.3 to 89.5 on COCONut-PanCap, showing that bridge tokens better bind text spans to image regions.
(2) \emph{The largest gains appear on long, dense captions.}
The gains of \model over baselines are strongest on COCONut-PanCap, which contains longer captions with dense object references, showing that explicit bridge-token prediction is more effective for challenging many-to-many text--image alignment.
(3) \emph{Pretrained VLM representations are essential for robust correspondence learning.}
The random-initialization gap highlights the value of pretrained visual and language representations for object recognition, referential understanding, and cross-modal alignment, especially in long, dense inputs.
(4) \emph{Detection-based grounding is not sufficient for full correspondence.}
Detector-based baselines remain competitive on mask-only or IoU metrics, but lag on joint correspondence, showing that the key challenge is identifying visually referential text spans and matching them to the image.
(5) \emph{\model handles large-vocabulary category lists in a single forward pass.}
On LVIS-minival, \model improves JointF1 from 23.1 to 29.9 over the strongest baseline, a 29\% relative gain.
While detector-based baselines require chunking the LVIS vocabulary into multiple inference passes, \model processes the full category list at once.
(6) \emph{Thresholded F1 and soft IoU reveal different strengths.}
Some detector-based baselines achieve strong soft IoU metrics, reflecting good partial localization, while \model achieves the best thresholded F1 metrics, showing better text--image correspondence decisions.

\subsection{Ablations}

\begin{table*}[t]
  \centering
  \small
\caption{Ablation study. We train each variant for 15K steps and report the average score for each metric across tasks. Ablations are performed sequentially; the vision encoder is frozen for the scoring-function, segmentation-head, and attention ablations, and fine-tuned for the subsequent vision-feature and bridge-assignment ablations. Bidirectional attention is used for the vision-feature and bridge-assignment ablations.}
\renewcommand{\arraystretch}{0.9}
\resizebox{\linewidth}{!}{
\begin{tabular}{ll*{6}{w{c}{1.3cm}}}
    \toprule
   \textbf{ Component} & \textbf{Variation} & {\bf JointF1}
    & {\bf MaskF1} & {\bf TextF1}
    & {\bf mJS}
    & {\bf mMaskIoU} & {\bf mSpanIoU} \\

% \midrule
% \multirow{2}{*}{\textbf{Data augmentation}}
%     & w/ augmentation        & \textbf{49.3} & \textbf{56.0} & \textbf{61.2} & 59.4 & 59.0 & 72.9 \\
%     & w/o augmentation       & 48.9 & 55.8 & 60.5 & \textbf{59.7} & \textbf{59.5} & \textbf{72.9} \\
    
\midrule
\multirow{2}{*}{\textbf{Scoring function}}
    & Dot product           & 48.7 & 55.4 & 60.3 & \textbf{59.9} & \textbf{59.7} & \textbf{73.3} \\
    & Bilinear              & \textbf{49.3} & \textbf{56.0} & \textbf{61.2} & 59.4 & 59.0 & 72.9 \\

\midrule
\multirow{4}{*}{\textbf{Image segmentation head}}
    & Bilinear-up (4x4)     & 47.8 & 54.3 & 60.3 & 59.1 & 58.0 & \textbf{73.3} \\
    & Two Conv-up (4x4)     & \textbf{49.3} & 56.0 & \textbf{61.2} & \textbf{59.4} & 59.0 & 72.9 \\
    & Three Conv-up (2x2)   & 48.8 & 56.3 & 60.1 & 59.0 & \textbf{59.5} & 71.6 \\
    & Four Conv-up (1x1)    & 48.8 & \textbf{56.4} & 60.6 & 58.5 & 59.3 & 70.4 \\

\midrule
\multirow{2}{*}{\textbf{Attention}}
    & Causal                & 49.3 & 56.0 & 61.2 & 59.4 & 59.0 & 72.9 \\
    & Bidirectional         & \textbf{51.3} & \textbf{57.2} & \textbf{62.6} & \textbf{61.4} & \textbf{60.4} & \textbf{74.7} \\

\midrule
\multirow{3}{*}{\textbf{Vision feature}}
    & Vision encoder feature + VLM vision tokens
                              & 52.4 & 59.4 & 62.5 & \textbf{62.2} & \textbf{62.7} & \textbf{73.9} \\
    & Vision encoder feature only
                              & 51.8 & 59.1 & 61.9 & 61.7 & 62.7 & 73.4 \\
    & VLM vision tokens only  & \textbf{52.5} & \textbf{59.5} & \textbf{62.8} & 61.0 & 61.4 & 72.8 \\

\midrule
\multirow{4}{*}{\textbf{Bridge token assignment}}
    & Raster order (64 tokens)
                                & 43.4 & 53.3 & 58.4 & 48.3 & 51.6 & 60.6 \\
    & Spatial assignment (8x8 tokens)
                                & 51.7 & 59.2 & 62.5 & 60.8 & 61.7 & 72.4 \\
    & Spatial assignment (11x11 tokens)
                                & 51.3 & 59.1 & 61.4 & 61.3 & 62.5 & 72.6 \\
    & Multi-scale spatial assignment
                                & \textbf{52.4} & \textbf{59.4} & \textbf{62.5} & \textbf{62.2} & \textbf{62.7} & \textbf{73.9} \\

    \bottomrule
  \end{tabular}
  }
  \label{tab:ablation}
\vspace{-5pt}
\end{table*}

\begin{figure}[t!]
    \centering
\includegraphics[width=\textwidth]{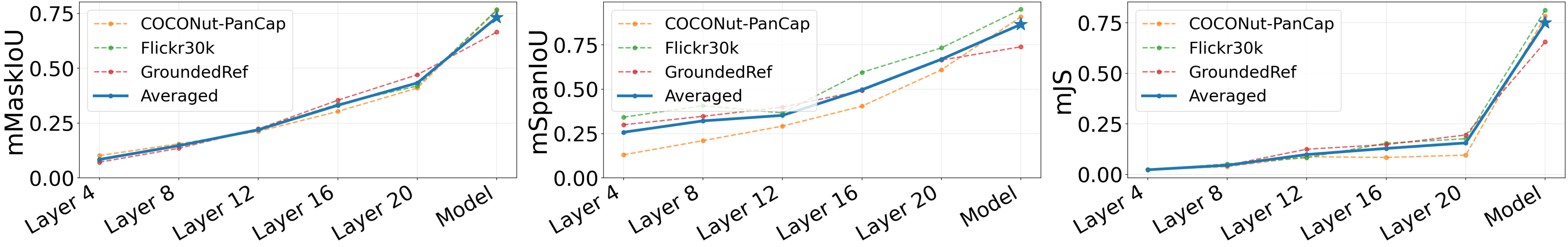}
\caption{
\textbf{Grounding emerges progressively across language model layers.}
We probe bridge-to-text and bridge-to-vision attention across language-model layers and evaluate the induced masks with mSpanIoU, mMaskIoU, and mJS.
Grounding quality improves progressively across layers, and the final prediction heads provide a further substantial boost.
}
\vspace{-10pt}
    \label{fig:layeriou}
\end{figure}

We ablate the main architectural choices of \model. 
Each variant is trained for 15K steps, and we report the average performance across evaluation tasks in Tab.~\ref{tab:ablation}.  

% \textbf{Data augmentation.}
% We first study the effect of data augmentation.
% It slightly improves the thresholded metrics, increasing JointF1 from 48.9 to 49.3, MaskF1 from 55.8 to 56.0, and TextF1 from 60.5 to 61.2.
% The soft IoU-based metrics are comparable, with the non-augmented model showing slightly higher mJS and mMaskIoU.
% Overall, data augmentation provides a small but consistent gain on the primary F1 metrics, so we use it by default.

\textbf{Scoring function.}
We compare dot-product and bilinear scoring for predicting image and text masks from bridge-token features.
Bilinear scoring improves the thresholded F1 metrics, suggesting that the learned metric compatibility separates foreground from background tokens.
Dot-product scoring performs slightly better on IoU-based metrics, indicating that simpler similarity may produce smoother or better-calibrated continuous masks.

\textbf{Image segmentation head.} % PASS
We ablate the image segmentation decoder design.
Bilinear upsampling performs worse than convolutional upsampling.
Among convolutional variants, the two-layer decoder gives the best JointF1, TextF1, and mJS, while deeper decoders slightly improve MaskF1 or mMaskIoU but reduce text and joint correspondence quality.
This suggests that increasing decoder depth can improve mask-only localization, but may not improve the alignment between text and image segments.
We therefore use the two-convolution upsampling head by default.

\textbf{Attention pattern.} %pass
We ablate the attention pattern in the LLM full-attention layers.
Replacing causal attention with bidirectional attention consistently improves performance across metrics.
This suggests that correspondence prediction benefits from full image--text--bridge context, since the task requires resolving all alignments jointly rather than generating tokens autoregressively.

\textbf{Vision features.} 
We compare two sources of vision features for the image segmentation head: vision encoder features and VLM vision tokens.
Using only VLM vision tokens yields the best F1 metrics, suggesting that the multimodal tokens provide stronger semantic context for deciding correspondences.
In contrast, combining vision encoder features with VLM vision tokens gives the best soft localization metrics, indicating that the original vision encoder features preserve fine-grained spatial detail useful for mask quality.
We therefore use the fused representation by default, as it improves IoU-based correspondence and localization metrics while remaining competitive on thresholded F1 metrics.

\textbf{Bridge token assignment.} % PASS
% Finally, we ablate how ground-truth correspondences are assigned to bridge tokens during training.
% A simple raster-order assignment with 64 tokens performs substantially worse, achieving only 43.4 JointF1 and 48.3 mJS.
% Replacing it with spatial assignment over an $8\times8$ or $11\times11$ grid improves JointF1 to 51.7 and 51.3, respectively, showing that spatial priors are important for learning stable correspondence queries.
% Our multi-scale spatial assignment performs best overall, reaching 52.4 JointF1, 59.4 MaskF1, 62.5 TextF1, 62.2 mJS, and 62.7 mMaskIoU.
% This confirms that assigning bridge tokens across multiple spatial scales helps the model cover objects with different sizes and locations, and we use this assignment strategy in the final model.
We ablate how ground-truth correspondences are assigned to bridge tokens during training.
A raster-order assignment performs substantially worse, while spatial grid-based assignment yields clear gains, indicating that localized priors are important for learning stable correspondence queries.
Our multi-scale spatial assignment performs best overall, showing that bridge tokens benefit from covering objects at different sizes and locations.
% We therefore use multi-scale spatial assignment in the final model.

\begin{figure}[t!]
    \centering
\includegraphics[width=\textwidth]{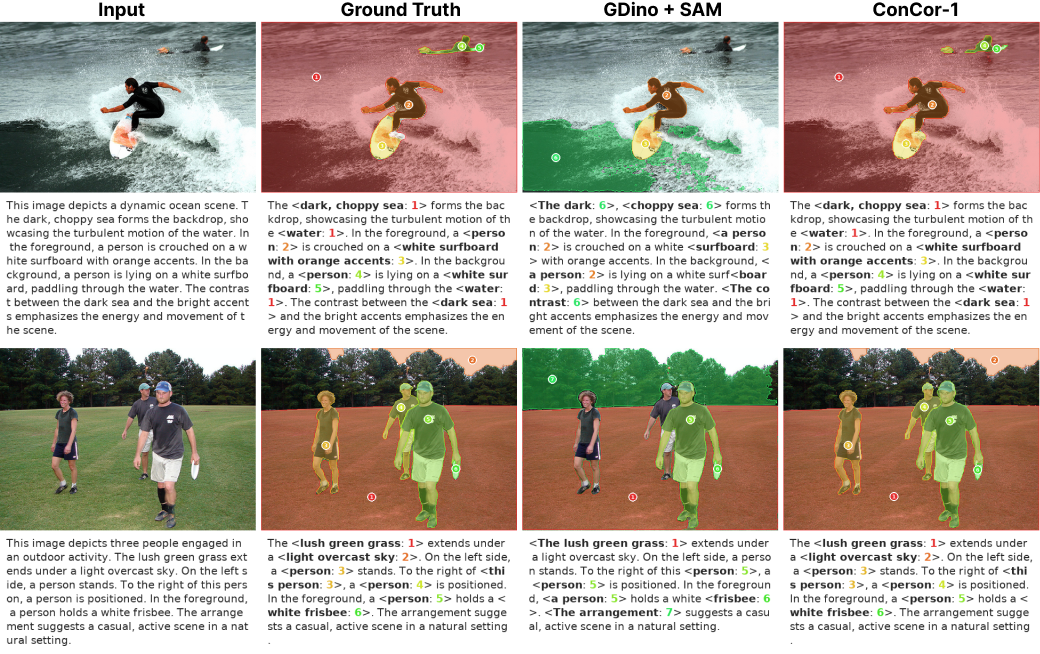}
    \caption{
\textbf{Qualitative comparison on COCONut-PanCap.}
We visualize ground truth, GDINO+SAM, and \model predictions, using matching colors for grounded text spans and image masks.
\model better disambiguates referred instances and recovers more complete text--image correspondences, while GDINO+SAM can confuse different people or objects mentioned in the caption and may miss some grounded entities. Additional examples are provided in Appendix~\ref{fig:examples-compare-app}.
}
    \label{fig:vis-compare}
\vspace{-10pt}
\end{figure}

\subsection{Analysis and visualizations}

\textbf{Layer-wise grounding analysis.}
We further analyze where grounding information appears inside the VLM by directly using the attention between bridge tokens and text or vision tokens as a prediction signal.
Specifically, for each selected language-model layer, we treat bridge-to-text attention as a text-mask prediction and bridge-to-vision attention as an image-mask prediction, and evaluate the resulting correspondences on COCONut-PanCap, Flickr30k, and GroundedRef.
As shown in Fig.~\ref{fig:layeriou}, lower layers contain little usable grounding information, while later layers yield consistently higher mMaskIoU, mSpanIoU, and mJS.
This trend suggests that bridge tokens progressively accumulate cross-modal correspondence information as they pass through the language model.
Nevertheless, the final model substantially outperforms all attention-only probes, showing that the learned segmentation heads are crucial for producing precise text spans and image masks.

\textbf{Qualitative comparison.}
Fig.~\ref{fig:vis-compare} shows qualitative examples on COCONut-PanCap.
GDINO+SAM often detects plausible regions but struggles to use the caption context to determine which instance is being referred to, especially when multiple people or similar objects appear in the image.
By jointly predicting text masks, image masks, and their correspondences, \model better disambiguates referred instances and produces more complete text--image alignments.

\textbf{Text-to-image attention analysis and visualization.}

We analyze text-to-image attention at full-attention Layers 4, 8, 12, 16, and 20 on the image-caption evaluation datasets. For each phrase, we average attention over heads and constituent text tokens, and compare the resulting visual-token map with its ground-truth mask. As shown in Tab.~\ref{tab:t2v_attn}, \model consistently achieves higher attention IoU and top-1 hit rate and lower spatial entropy than Qwen3.5 and Qwen3.5-FT across all analyzed layers, indicating more spatially aligned and concentrated phrase-level attention.

\begin{table}[h]
\centering
\scriptsize
\setlength{\tabcolsep}{3.5pt}
\renewcommand{\arraystretch}{1.05}
\caption{Quantitative text-to-image attention analysis across full-attention layers, averaged over the image-caption evaluation datasets. Attn-IoU measures the overlap between the highest-attention visual tokens and the phrase-specific ground-truth mask, using the ground-truth mask area to determine the number of selected tokens. Top-1 Hit measures whether the visual token receiving the maximum attention lies inside the ground-truth mask. Entropy measures the normalized spatial dispersion of attention, where lower values indicate more concentrated attention.}
\label{tab:t2v_attn}
\begin{tabular}{llccccc}
\toprule
\textbf{Metric} & \textbf{Model} &
\textbf{L4} & \textbf{L8} & \textbf{L12} &
\textbf{L16} & \textbf{L20} \\
\midrule
Attn-IoU $\uparrow$
& Qwen3.5 & 0.16 & 0.20 & 0.22 & 0.22 & 0.21 \\
& Qwen3.5-FT & 0.16 & 0.27 & 0.23 & 0.19 & 0.15 \\
& \textbf{\model} & \textbf{0.29} & \textbf{0.42} &
\textbf{0.48} & \textbf{0.33} & \textbf{0.57} \\
\midrule
Top-1 Hit $\uparrow$
& Qwen3.5 & 0.32 & 0.43 & 0.43 & 0.43 & 0.40 \\
& Qwen3.5-FT & 0.41 & 0.56 & 0.47 & 0.36 & 0.27 \\
& \textbf{\model} & \textbf{0.56} & \textbf{0.70} &
\textbf{0.70} & \textbf{0.55} & \textbf{0.75} \\
\midrule
Entropy $\downarrow$
& Qwen3.5 & 0.82 & 0.69 & 0.73 & 0.74 & 0.73 \\
& Qwen3.5-FT & 0.78 & 0.73 & 0.71 & 0.70 & 0.63 \\
& \textbf{\model} & \textbf{0.73} & \textbf{0.66} &
\textbf{0.65} & \textbf{0.68} & \textbf{0.54} \\
\bottomrule
\end{tabular}
\end{table}

\begin{figure}[t!]
    \centering
\includegraphics[width=0.7\textwidth]{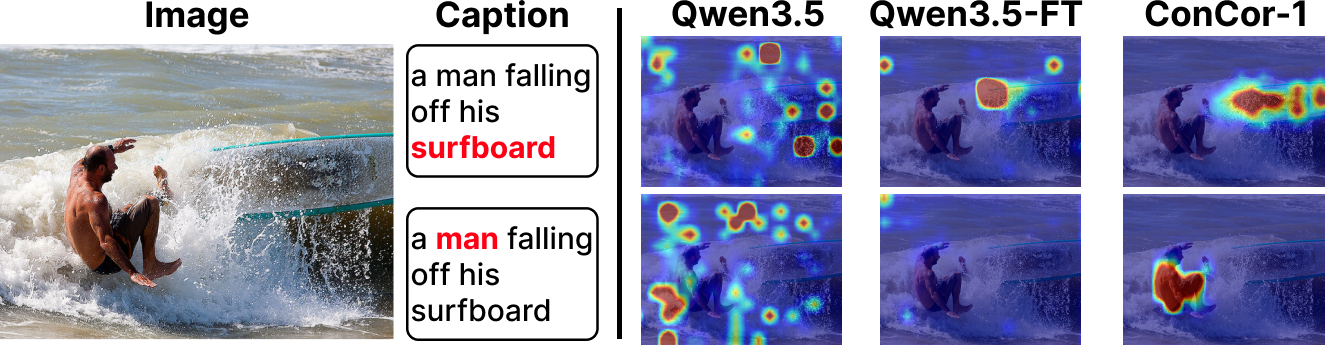}
    \caption{
\textbf{Text-to-image attention visualization.}
    We visualize layer-20 attention from selected text tokens (highlighted in red) to image tokens for the original Qwen3.5, Qwen3.5 fine-tuned on our dataset, and our model.
    Compared with the diffuse or noisy baselines, \model produces more localized and semantically aligned attention. Additional visualizations are provided in Appendix~\ref{fig:t2v-attention-app}.
}
    \label{fig:t2v-att-compare}
\end{figure}

We visualize Layer 20 in Fig.~\ref{fig:t2v-att-compare} because it provides the strongest alignment for \model, achieving the highest attention IoU and top-1 hit rate and the lowest entropy among the analyzed layers. The original Qwen3.5 exhibits scattered attention patterns, with high responses on unrelated background regions and objects.
Fine-tuning Qwen3.5 on our data makes the attention more localized, but the attended regions are still noisy and do not consistently cover the correct entity.
In contrast, our model produces more compact and semantically aligned attention maps: the token ``surfboard'' attends to the surfboard region, while the token ``man'' attends to the person.
This indicates that our explicit bridge-token correspondence objective not only improves final grounding predictions, but also induces cleaner cross-modal alignment in the underlying VLM representations.

\begin{figure}[t!]
    \centering
\includegraphics[width=\textwidth]{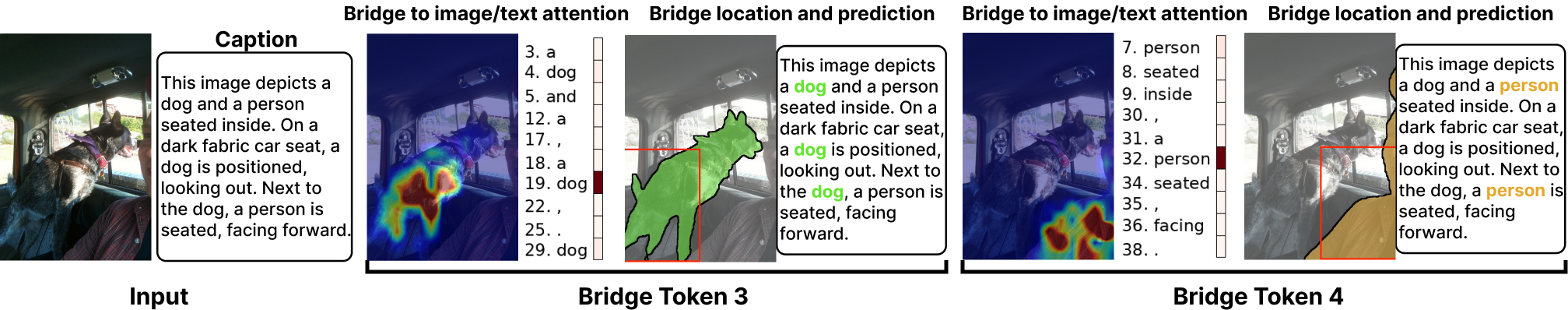}
\caption{
\textbf{Bridge-token attention and prediction.}
We visualize two bridge tokens at layer 20 for one test sample, including bridge-to-image attention, bridge-to-text attention, spatial prior (red bounding box), and final prediction.
Different bridge tokens specialize to distinct object-level correspondence hypotheses, such as the dog and the seated person.
Additional examples are provided in Appendix~\ref{fig:bridge-att-app}.
}
\vspace{-10pt}
    \label{fig:bridge-att}
\end{figure}

\textbf{Bridge-token attention visualization.}
We visualize bridge-to-image and bridge-to-text attention at layer 20, together with the spatial prior and final prediction associated with each bridge token.
As shown in Fig.~\ref{fig:bridge-att}, different bridge tokens attend to different visual regions and text mentions, forming distinct correspondence hypotheses.
For example, one bridge token focuses on the dog region and attends to repeated mentions of ``dog'', while another focuses on the seated person and attends to the corresponding person mentions.
The final predictions show that these bridge-specific attention patterns are converted into coherent text--image correspondences, supporting the role of bridge tokens as object-level hypotheses that bind language and vision.

\section{Discussion and Future Work}

We present bidirectional concept correspondence as a structured formulation for grounding text segments, image segments, and their alignments from an image--text pair. This formulation unifies caption grounding, referring-expression grounding, and open-vocabulary grounding under a single output space.
Our model \model provides a first step toward this goal by explicitly predicting text masks, image masks, and their correspondences.
Future work includes extending correspondence to video, where alignments must persist over time, and using it as a training objective for better-aligned vision-language models. This task can also support the evaluation and improvement of vision-language models, such as checking whether generated captions are visually grounded or whether text-to-image outputs faithfully realize prompt entities and relations. 

\section*{Acknowledgement}

This project was partially funded by Toyota Motor Inc.

% \section{Discussion and Future Work}

% We present bidirectional concept correspondence as a structured formulation for grounding text segments, image segments, and their alignments from an image--text pair. This formulation provides a structured way to study whether multimodal models truly connect linguistic concepts with visual evidence and can unify caption grounding, referring-expression grounding, and category-based segmentation under a single output space.
% Future work includes extending this formulation to video, where correspondences must be maintained across time. Another direction is to use concept correspondence as a pretraining objective for learning fine-grained cross-modal alignment. Finally, this task can support the evaluation and improvement of other vision-language models, such as checking whether generated image captions are visually grounded or whether text-to-image outputs faithfully realize the entities and relations in a prompt.

% \clearpage
{
\small
\bibliographystyle{unsrt}
\bibliography{main}
}

%%%%%%%%%%%%%%%%%%%%%%%%%%%%%%%%%%%%%%%%%%%%%%%%%%%%%%%%%%%%

\appendix

\newpage

\section{Limitations and Societal Impact}
\paragraph{Limitations.}
First, \model's mask quality remains behind that of specialized detection and segmentation models. This is partly because our image-mask head is intentionally lightweight and designed for joint text-image correspondence prediction rather than optimizing segmentation quality alone. Second, our approach depends on correspondence-style supervision, where grounded text spans and image masks are jointly annotated. Since most existing datasets are not natively annotated in this format, we convert several grounded-caption, detection, segmentation, and visual-coreference datasets into a unified representation, which may introduce noise and leave some forms of supervision underused. Future work could study whether large-scale single-modality datasets, such as segmentation-only or text-only coreference data, can further improve the visual or linguistic side of correspondence prediction. Third, while our evaluation covers captions, referring expressions, and category-name inputs, future benchmarks should include longer and more compositionally complex captions with more entities, repeated mentions, nested references, attributes, relations, and discourse-level coreference. These settings would better test whether models can recover complete correspondence structures under realistic language complexity.

\paragraph{Societal Impact.}
The societal impact of this work depends on how grounding models are used in downstream systems. More precise image-text correspondence can improve interpretability by making vision-language predictions spatially attributable, which may benefit applications such as visual search, dataset annotation, assistive perception, and embodied agents. However, grounding errors may also propagate to downstream decisions, especially in human-facing or safety-critical settings. For example, incorrect correspondences could cause an assistive system to refer to the wrong object, a robotic system to act on an unintended target, or an annotation pipeline to amplify dataset bias. Since grounding models reflect the distributions and biases of their training data, they may also perform unevenly across object categories, visual domains, languages, and cultural contexts. We therefore view this work as a step toward more transparent and structured vision-language understanding, but recommend careful validation before deployment in high-stakes applications.

\section{Implementation Details}
\label{app:implementation_details}
The implementation of \model follows a grounding-only design: the model does not autoregressively decode an output string, but instead appends learnable bridge tokens to the multimodal input sequence and directly predicts image masks, text masks, and correspondence presence from the final token representations.

\subsection{Backbone and Input Sequence}

We instantiate the backbone from the pretrained Qwen3.5-0.8B vision-language model. 
% The parameter comparision between \model and its Qwen3.50.8B backbone is in Tab.~\ref{tab:param_comparison}. 
\model uses the VLM backbone as a feature extractor to obtain the final hidden state sequence, rather than as a conditional language generator. We therefore apply bidirectional attention to the full-attention layers of the Qwen3.5 language backbone, allowing the bridge tokens to access the complete multimodal context and to differentiate from one another, while preserving the original attention behavior in the linear-attention layers. This preserves the pretrained hybrid-attention structure of Qwen3.5, but allows the full-attention layers to jointly contextualize visual tokens, text tokens, and bridge tokens.

For each image--text pair, we construct a flat multimodal sequence:
\[
[\texttt{<vision\_start>},\ \texttt{<image\_pad>}_{1:N_v},\ \texttt{<vision\_end>},\ x^{\text{text}}_{1:N_t},\ b_{1:Q}],
\]
where \(N_v\) is the number of merged visual tokens, \(N_t\) is the number of text tokens, and \(Q\) is the number of bridge tokens. 

For all main-table experiments, inspired by detection-based grounding methods that operate under a controlled image scale, we fix the image pixel budget at 1,003,520. We retain Qwen3.5's native aspect-ratio-preserving dynamic resizing, such that the image height and width adapt to the original aspect ratio rather than being forced to a fixed square resolution. This provides an approximately fixed visual-token budget across images while preserving their original geometry.

\subsection{Bridge Tokens and Multi-scale Assignment}

The bridge tokens are learnable embeddings appended after the image and text tokens. Each bridge token represents one candidate text--image correspondence. We organize bridge tokens into multi-scale spatial groups. Specifically, we use grid levels from \(1\times1\) to \(10\times10\). A level \(s\) contributes \(s^2\) bridge tokens, so the total number of bridges is
\[
Q=\sum_{s=1}^{10} s^2 = 385.
\]
The bridge tokens are ordered by scale, and within each scale they are ordered in raster order over the corresponding grid. This gives the bridge set an implicit spatial and scale structure: coarse levels can represent large instances, while fine levels can represent smaller localized instances.

During training, each ground-truth correspondence is assigned to one bridge token. For each bridge token, we define its associated grid cell according to its scale and spatial position. We then perform Hungarian matching between masks of each correspondence and bridge tokens, using the mask--cell overlap as the primary criterion and the distance between the mask centroid and the cell center as a tie-breaker. 

\subsection{Prediction Heads}

Given the final hidden states from the backbone, \model extracts three groups of token features: bridge-token features, text-token features, and visual-token features. The bridge-token features are shared by all three prediction heads. Each head first produces logits and then applies a sigmoid function to obtain probabilities for binary prediction.

\paragraph{Presence head.}
The presence head predicts whether a bridge token corresponds to a valid grounded correspondence. It takes the final hidden state of each bridge token as input and outputs a scalar logit:
\[
z^{\mathrm{pres}}_j = f_{\mathrm{pres}}(\mathbf{b}_j),
\]
where \(\mathbf{b}_j\in\mathbb{R}^{1024}\) denotes the representation of the \(j\)-th bridge token. We implement \(f_{\mathrm{pres}}\) as a SwiGLU MLP following the gated feed-forward style of the backbone: a gated projection and an up projection are combined with a SiLU nonlinearity, normalized, and projected to a scalar. The logit is then converted into a presence probability by a sigmoid function. The resulting presence score is used to determine whether the bridge should produce a valid text--image correspondence.

\paragraph{Text segmentation head.}
The text segmentation head predicts the text span associated with each bridge token. Rather than predicting explicit start and end indices, we formulate text grounding as token-level binary segmentation. For a bridge representation \(\mathbf{b}_j\) and a text-token representation \(\mathbf{x}^{\mathrm{text}}_\ell\), we project both into a shared correspondence space using independent SwiGLU MLPs:
\[
\phi_b(\mathbf{b}_j),\ \phi_t(\mathbf{x}^{\mathrm{text}}_\ell)\in\mathbb{R}^{256}.
\]
We then compute a learned bilinear compatibility score:
\[
z^{\mathrm{text}}_{j,\ell}
=
\phi_t(\mathbf{x}^{\mathrm{text}}_\ell)^\top
W_{\mathrm{text}}
\phi_b(\mathbf{b}_j).
\]
Applying this scorer to all bridge--text-token pairs yields
\[
\mathbf{Z}^{\mathrm{text}}\in\mathbb{R}^{B\times Q\times N_t}.
\]
\paragraph{Vision segmentation head.}
The vision segmentation head predicts the image segment associated with each bridge token. It contains two components: a feature decoder that reconstructs dense visual features from the compressed VLM visual tokens, and a mask predictor that scores each bridge against the decoded visual features.

The Qwen3.5 vision encoder uses \(16\times16\) image patches followed by \(2\times2\) spatial merging. Thus, each merged visual token corresponds to a \(32\times32\) image region. Directly predicting masks over these merged tokens is spatially coarse, so our model uses a fine decoding path. First, each merged visual token is expanded into four patch-level features by four individual MLPs, corresponding to the top-left, top-right, bottom-left, and bottom-right \(16\times16\) patches before spatial merging. This step recovers the local \(2\times2\) patch layout inside each merged token.
To preserve low-level visual detail, we additionally use a shortcut from the vision encoder to fuse the decoded patch-level features with the corresponding pre-merger ViT patch features. This fusion combines the language-model contextual representation with local visual features that are closer to the original image grid.

After this patch-level reconstruction, the features are rearranged into a spatial feature map and refined by a lightweight convolutional upsampling decoder. The decoder contains two successive upsampling blocks. Each block applies a stride-2 transposed convolution, followed by a SiLU activation, a depthwise \(3\times3\) convolution for local refinement, and a two-dimensional layer normalization. Since each block upsamples the feature map by a factor of two, the two-block decoder refines the representation from the \(16\times16\) patch level to a \(4\times4\) pixel-block grid.

Given a decoded visual feature \(\mathbf{v}_n\) at spatial location \(n\) and a bridge representation \(\mathbf{b}_j\), the mask predictor projects both into a shared \(256\)-dimensional correspondence space and applies a learned bilinear scorer:
\[
z^{\mathrm{vis}}_{j,n}
=
\phi_v(\mathbf{v}_n)^\top
W_{\mathrm{vis}}
\phi_b(\mathbf{b}_j).
\]
This produces a fine-resolution mask logit map for every bridge token:
\[
\mathbf{Z}^{\mathrm{vis}}\in\mathbb{R}^{B\times Q\times N_{\mathrm{fine}}}.
\]
We supervise this fine mask prediction directly during training. At inference time, the predicted fine mask grid is reshaped into its spatial layout and bilinearly interpolated to the original image resolution before thresholding and evaluation.

\subsection{Training Loss}

We supervise the three outputs of each bridge token with a presence loss, a text segmentation loss, and a visual segmentation loss. Expanding the visual term into its sampled-point BCE and Dice components, the total objective is a weighted sum of four terms:
\[
\mathcal{L}
=
\lambda_{\mathrm{pres}}\mathcal{L}_{\mathrm{pres}}
+
\lambda_{\mathrm{text}}\mathcal{L}_{\mathrm{text}}
+
\underbrace{
\lambda_{\mathrm{bce}}\mathcal{L}_{\mathrm{bce}}^{\mathrm{pt}}
+
\lambda_{\mathrm{dice}}\mathcal{L}_{\mathrm{dice}}^{\mathrm{pt}}
}_{\mathcal{L}_{\mathrm{vis}}}.
\]
We use
\(\lambda_{\mathrm{text}}=1.0\),
\(\lambda_{\mathrm{bce}}=2.0\),
\(\lambda_{\mathrm{dice}}=0.5\),
and \(\lambda_{\mathrm{pres}}=2.0\)
for all main experiments; this setting is selected by the sweep reported in Tab.~\ref{tab:loss-weight-ablation}.

\paragraph{Presence loss.}
The bridge assignment defines the binary presence target. A bridge token is labeled positive if it is matched to a ground-truth correspondence and negative otherwise. We train the presence head with binary cross-entropy to select which candidate bridge tokens correspond to valid grounded concepts. This term is scaled by \(\lambda_{\mathrm{pres}}\) and, unlike the two segmentation terms, is applied to every bridge token rather than only the matched positives.

\paragraph{Text segmentation loss.}
The text target for each positive bridge is a binary mask over text tokens. We derive this mask from the ground-truth character spans: a text token is labeled positive if its offset overlaps the character span of the assigned correspondence. The text segmentation loss is binary cross-entropy over valid text-token positions, scaled by \(\lambda_{\mathrm{text}}\).

\paragraph{Visual segmentation loss.}
The visual target for each positive bridge is a binary ground-truth mask at the processed image resolution. We supervise visual masks with a PointRend-style~\cite{kirillov2020pointrend} point-sampled loss. For each matched positive bridge, we reshape its finest visual logits into a spatial logit map, over-sample candidate points uniformly, and select a fixed fraction of points with the highest uncertainty, measured by the negative absolute logit magnitude. The remaining points are sampled uniformly for spatial coverage. The same normalized coordinates are used to sample both predicted logits and high-resolution binary ground-truth masks.

As written above, the visual loss \(\mathcal{L}_{\mathrm{vis}}\) is the weighted sum of a sampled-point BCE term and a sampled-point Dice term, with weights \(\lambda_{\mathrm{bce}}\) and \(\lambda_{\mathrm{dice}}\). The BCE term provides local point-wise supervision, while the Dice term encourages mask-level overlap. Visual supervision is applied only to matched positive bridge tokens; unmatched bridges are trained only through the presence loss.

\paragraph{Choice of loss weights.}
The four terms are not naturally on the same scale, so their relative weighting matters. We sweep \((\lambda_{\mathrm{text}},\lambda_{\mathrm{bce}},\lambda_{\mathrm{dice}},\lambda_{\mathrm{pres}})\) over five settings and report the results in Tab.~\ref{tab:loss-weight-ablation}. All runs in this sweep share the same configuration: a jointly fine-tuned vision encoder, bidirectional attention, and \(30\mathrm{K}\) training steps. Each entry is averaged over our evaluation benchmarks, and we report the six headline metrics. \((1,2,0.5,2)\) is best on all six metrics, so we adopt it for the main experiments.

\begin{table}[hb]
  \centering
  \scriptsize
  \setlength{\tabcolsep}{4pt}
  \renewcommand{\arraystretch}{1}
  \caption{
  Effect of the loss weights \((\lambda_{\mathrm{text}},\lambda_{\mathrm{bce}},\lambda_{\mathrm{dice}},\lambda_{\mathrm{pres}})\).
  % Each number is averaged over our evaluation benchmarks except EntitySeg.
  }
  \label{tab:loss-weight-ablation}
  \begin{tabular}{@{}lcccccc@{}}
    \toprule
    \textbf{\(\lambda_{\mathrm{text}}\):\(\lambda_{\mathrm{bce}}\):\(\lambda_{\mathrm{dice}}\):\(\lambda_{\mathrm{pres}}\)}
    & \textbf{JointF1} & \textbf{MaskF1} & \textbf{TextF1} & \textbf{mJS} & \textbf{mMaskIoU} & \textbf{mSpanIoU} \\
    \midrule
    1 : 0.5 : 0.5 : 1
      & 60.5 & 66.8 & 70.7 & 66.8 & 66.5 & 78.4 \\
    1 : 2 : 0.5 : 1
      & 61.3 & 67.2 & 71.3 & 67.3 & 66.8 & 78.9 \\
    1 : 5 : 5 : 1
      & 59.9 & 66.1 & 69.9 & 66.0 & 66.0 & 77.4 \\
    1 : 2.5 : 0 : 2
      & 59.7 & 65.5 & 70.8 & 66.1 & 65.6 & 79.3 \\
    \textbf{1 : 2 : 0.5 : 2}
      & \textbf{62.0} & \textbf{67.7} & \textbf{72.1} & \textbf{68.0} & \textbf{67.3} & \textbf{79.5} \\
    \bottomrule
  \end{tabular}
\end{table}

\subsection{Backbone and Input Sequence}

% \begin{table}[b]
% \centering
% \scriptsize
% \setlength{\tabcolsep}{5pt}
% \renewcommand{\arraystretch}{1}
% \caption{Parameter comparison between the Qwen3.5-0.8B backbone and \model.}
% \label{tab:param_comparison}
% \begin{tabular}{lccc}
% \toprule
% \textbf{Model / Component} & \textbf{Params (M)} & \textbf{Added Params (M)} & \textbf{Relative Size} \\
% \midrule
% Qwen3.5-0.8B backbone & 852.99 & -- & 1.000$\times$ \\
% \midrule
% \quad Presence head & 0.52 & 0.52 & -- \\
% \quad Text segmentation head & 1.11 & 1.11 & -- \\
% \quad Visual segmentation head & 12.01 & 12.01 & -- \\
% \quad New bridge embeddings & 0.145 & 0.145 & -- \\
% \midrule
% \textbf{\model} & \textbf{866.79} & \textbf{13.80} & \textbf{1.016$\times$ (+1.62\%)} \\
% \bottomrule
% \end{tabular}
% \end{table}

\begin{table}[b]
\centering
\scriptsize
\setlength{\tabcolsep}{3.5pt}
\renewcommand{\arraystretch}{1.05}
\caption{Model size and computational profile. FLOPs, peak GPU memory,
and latency are measured for one forward pass with a $512{\times}512$
image, 512 text tokens, batch size 1, bf16 precision, and an A100-80GB GPU.}
\label{tab:efficiency_comparison}
\begin{tabular}{lccccc}
\toprule
\textbf{Model}
& \textbf{Seq. Len.}
& \textbf{FLOPs}
& \textbf{Peak Mem.}
& \textbf{Latency}
& \textbf{Params} \\
\midrule
Qwen3.5-0.8B backbone
& 770
& 1.061 T
& 1.687 GiB
& 54.0 ms
& 852.99 M \\
\model
& 1,155
& 1.549 T
& 1.846 GiB
& 55.4 ms
& 866.79 M \\
\midrule
$\Delta$
& +385
& +46.0\%
& +9.5\%
& +2.6\%
& +1.62\% \\
\bottomrule
\end{tabular}
\end{table}

\begin{table}[h]
\centering
\scriptsize
\setlength{\tabcolsep}{4pt}
\renewcommand{\arraystretch}{1.05}
\caption{Component-level parameter and FLOPs breakdown. The computation induced by the bridge tokens is included in the language-model row.}
\label{tab:param_comparison_details}
\begin{tabular}{lrrrr}
\toprule
\textbf{Component} &
\textbf{Added Params} &
\multicolumn{3}{c}{\textbf{GFLOPs}} \\
\cmidrule(lr){3-5}
& \textbf{(M)} & \textbf{Backbone} & \textbf{\model} & $\boldsymbol{\Delta}$ \\
\midrule
Vision tower & -- & 221.5 & 221.5 & -- \\
Language model & -- & 839.3 & 1,279.2 & +439.9 \\
New bridge embeddings & 0.145 & -- & -- & -- \\
Visual segmentation head & 12.01 & -- & 46.3 & +46.3 \\
Text segmentation head & 1.11 & -- & 1.1 & +1.1 \\
Presence head & 0.52 & -- & 0.4 & +0.4 \\
\midrule
\textbf{Total} & \textbf{13.80} &
\textbf{1,060.8} & \textbf{1,548.5} & \textbf{+487.7} \\
\bottomrule
\end{tabular}
\end{table}

Tab.~\ref{tab:efficiency_comparison} compares the model size and computational profile of \model with its Qwen3.5-0.8B backbone. We profile one forward pass using a $512{\times}512$ image, 512 text tokens, batch size 1, bf16 precision, and an A100-80GB GPU. \model introduces 13.80M additional parameters, increasing the model size by 1.62\%. Its 385 bridge tokens increase the sequence length from 770 to 1,155, resulting in a 46.0\% increase in FLOPs. The measured peak GPU memory and forward latency increase by 9.5\% and 2.6\%, respectively.

Tab.~\ref{tab:param_comparison_details} further decomposes the parameter and computational overhead. Most of the additional computation comes from processing the bridge tokens through the 24-layer language backbone, which contributes an additional 439.9 GFLOPs. The three prediction heads together add 47.8 GFLOPs, with the visual segmentation head accounting for most of this cost.

\begin{table}[t]
  \centering
  \scriptsize
  \setlength{\tabcolsep}{4pt}
  \renewcommand{\arraystretch}{0.92}
  \caption{
  Training data mixture used for \model.
  We report the source-level mixture weight in the final training setup.
  The image augmentation column indicates the fraction of each source group sampled with collage or crop augmentation; the remaining fraction is sampled without image augmentation.
  }
  \label{tab:full-data-mixture}
  \begin{tabular}{@{}lcc@{}}
    \toprule
    {\bf Source dataset}
    & {\bf Mix weight}
    & {\bf Image aug.} \\
    \midrule

    \multicolumn{3}{c}{\emph{Instance Segmentation Data}} \\
    \midrule
    COCO~\cite{lin2015microsoftcococommonobjects}
      & 15\% & 80\% collage \\
    EntitySeg~\cite{qi2023highquality}
      & 10\% & 10\% crop \\
    PixMo Points~\cite{molmov1}
      & 10\% & 10\% crop \\
    COCONut~\cite{deng2025coconut}
      & 5\% & 70\% collage, 10\% crop \\
    SA-1B~\cite{kirillov2023segment}
      & 5\% & 10\% crop \\
    ADE20K~\cite{zhou2019semantic}
      & 5\% & 50\% collage, 20\% crop \\
    Roboflow-VL-100~\cite{robicheaux2025roboflow100}
      & 5\% & \textemdash \\
    \midrule

    \multicolumn{3}{c}{\emph{Caption Grounding Data}} \\
    \midrule
    GoldG~\cite{liu2023grounding}
      & 18\% & \textemdash \\
    COCONut-PanCap~\cite{deng2025coconut}
      & 15\% & \textemdash \\
    GroundedRef
      & 12\% & \textemdash \\
    \midrule

    Total
      & 100\% & \textemdash \\
    \bottomrule
  \end{tabular}
\end{table}

\subsection{Data Augmentation and Data Mixture}
We apply two image-level augmentations to improve the coverage of object layout and scale. The collage augmentation combines multiple training images into a single canvas using either a grid or horizontal/vertical concatenation. The corresponding instance masks are transformed into the collage coordinate system and re-encoded, and the text supervision is rebuilt according to the categories present in the augmented image. This increases the number of grounded instances per sample and distributes instances across different spatial cells, yielding denser supervision for the multi-scale bridge layout. 
The crop augmentation samples a random crop with constrained scale and aspect ratio, transforms the masks into the cropped image, and keeps only instances that remain sufficiently visible. The category text and character-span annotations are then rebuilt for the remaining instances. This augmentation improves robustness to partial objects, scale variation, and local context changes. It also changes the relative position and extent of objects within the image, which helps reduce center and layout biases in the training data.

We report the training data mixture in Tab.~\ref{tab:full-data-mixture}.

% \begin{table}[t]
%   \centering
%   \scriptsize
%   \setlength{\tabcolsep}{4pt}
%   \renewcommand{\arraystretch}{0.92}
%   \caption{
%   Training data mixture used for \model.
%   We report the source-level mixture weight in the final training setup.
%   The image augmentation column indicates the fraction of each source group sampled with collage or crop augmentation.
%   }
%   \label{tab:full-data-mixture}
%   \begin{tabular}{@{}lcc@{}}
%     \toprule
%     {\bf Source dataset}
%     & {\bf Mix weight}
%     & {\bf Image aug.} \\
%     \midrule

%     \multicolumn{3}{c}{\emph{Instance Segmentation Data}} \\
%     \midrule
%     COCO~\cite{lin2015microsoftcococommonobjects}
%       & 15\% & 20\% collage \\
%     COCONut~\cite{deng2025coconut}
%       & 10\% & 50\% collage/crop \\
%     PixMo Points~\cite{molmov1}
%       & 10\% & 10\% crop \\
%     SA-1B~\cite{kirillov2023segment}
%       & 10\% & 10\% crop \\
%     ADE20K~\cite{zhou2019semantic}
%       & 5\% & 60\% collage/crop \\
%     Roboflow-VL-100~\cite{robicheaux2025roboflow100}
%       & 5\% & \textemdash \\
%     \midrule

%     \multicolumn{3}{c}{\emph{Caption Grounding Data}} \\
%     \midrule
%     GoldG~\cite{liu2023grounding}
%       & 18\% & \textemdash \\
%     COCONut-PanCap~\cite{deng2025coconut}
%       & 15\% & \textemdash \\
%     GroundedRef
%       & 12\% & \textemdash \\
%     \midrule

%     Total
%       & 100\% & \textemdash \\
%     \bottomrule
%   \end{tabular}
% \end{table}

\subsection{Inference}

During inference, \model produces a set of candidate correspondences from the bridge tokens. We keep bridge tokens whose presence scores exceed the inference threshold. For the main results, we use a presence threshold of \(0.1\).

For each retained bridge token, the text segmentation head outputs token-level logits over the input text. We apply sigmoid activation and threshold the resulting probabilities to obtain a binary text-token mask. We use a text threshold of \(0.45\) in the experiments.

For the visual prediction, the vision segmentation head outputs mask logits on the fine \(4\times4\) pixel-block grid. We reshape these logits into their spatial layout and bilinearly upsample them to the original image resolution. The upsampled logits are then passed through a sigmoid function to obtain pixel-level mask probabilities, which are thresholded to produce the final binary mask. We use a visual threshold of \(0.45\) for the main results.

Because overlapping bridge tokens may predict highly similar correspondences, we apply non-maximum suppression to the retained predictions. Predictions are ranked by their presence scores, and duplicate lower-scoring predictions are removed using an IoU threshold of \(0.5\). The remaining predictions are used as the final set of text--image correspondences.

\subsection{Ablation Details}
For the scoring-function and image-segmentation-head ablations, we use causal attention and keep the vision encoder frozen. We then ablate causal versus bidirectional attention while continuing to freeze the vision encoder. After selecting bidirectional attention, we jointly fine-tune the vision encoder for the vision-feature and bridge-token-assignment ablations. For all ablations, we use bridge levels \(1\times1\) to \(5\times5\) and \(8\times8\). The training mixture excludes EntitySeg, Roboflow-VL-100, and GoldG, while adding LVIS. Each model is trained for \(15\mathrm{K}\) steps, with an NMS IoU threshold of \(0.9\) at inference. All other settings follow the main experiments.

% For ablation experiments, we use bridge levels \(1\times1\) to \(5\times5\) and \(8\times8\). The training mixture excludes EntitySeg, Roboflow-VL-100 and GoldG, while adding LVIS. We train each model for \(15\mathrm{K}\) steps and use an NMS IoU threshold of \(0.9\) during inference. All other settings follow the main experiments.

\section{Evaluation metrics details}
\label{app:metrics}

Our evaluation framework assesses grounded caption predictions along three independent axes: \emph{text
segmentation} (whether the model correctly groups character spans), \emph{image segmentation} (whether the
predicted masks align with ground truth), and \emph{joint correspondence} (whether the model correctly associates
text spans with masks). Each axis follows the same general procedure: (1) compute a pairwise similarity matrix
between all predicted and ground-truth groups for a given image, (2) solve an optimal one-to-one assignment via
Hungarian matching, and (3) derive hard and soft scores from the matched pairs. All metrics are computed per image
and then macro-averaged across images.

\subsection{Notation}

Let $\mathcal{G} = \{G_1, \ldots, G_M\}$ denote the set of ground-truth groups for an image and
$\hat{\mathcal{G}} = \{\hat{G}_1, \ldots, \hat{G}_K\}$ the predicted groups. Each group $G_i$ consists of a set of
character spans $T_i \subseteq \mathbb{Z}$ (the union of all $[\text{start}, \text{end})$ intervals) and a binary
segmentation mask $S_i \in \{0,1\}^{H \times W}$. We write $M = |\mathcal{G}|$ for the number of GT groups and
$K = |\hat{\mathcal{G}}|$ for the number of predictions.

\subsection{Text Segmentation}

Text segmentation evaluates whether the model correctly identifies which character spans in the caption correspond
to each visual entity, without considering the predicted masks.

\paragraph{Span IoU.} For each pair $(G_i, \hat{G}_j)$, we compute the character-level Intersection over Union
between their span sets. All $[\text{start}, \text{end})$ intervals within a group are first unioned into a single
set of character indices:
\[
\text{spanIoU}(G_i, \hat{G}_j) = \frac{|T_i \cap \hat{T}_j|}{|T_i \cup \hat{T}_j|}
\]
This formulation correctly handles adjacent spans (e.g., $[0,5) + [5,10)$ versus $[0,10)$ yields IoU $= 1.0$) and
overlapping spans within a group.

\paragraph{Matching.} We construct an $M \times K$ similarity matrix $\mathbf{A}^{\text{text}}$ where
$A^{\text{text}}_{ij} = \text{spanIoU}(G_i, \hat{G}_j)$. Hungarian matching is applied by minimizing
$-\mathbf{A}^{\text{text}}$, producing an optimal one-to-one assignment $\pi$ between GT and predicted groups. No
similarity threshold is applied during matching; all assigned pairs are retained regardless of their score.

\paragraph{Hard metrics.} Given a threshold $\tau$, a matched pair $(G_i, \hat{G}_{\pi(i)})$ is counted as a true
positive if $\text{spanIoU}(G_i, \hat{G}_{\pi(i)}) \geq \tau$. Let $\mathrm{TP}_\tau$ denote the number of such
pairs. We define:
\begin{align}
\text{TextP}@\tau &= \frac{\mathrm{TP}_\tau}{K}, \quad
\text{TextR}@\tau = \frac{\mathrm{TP}_\tau}{M}, \quad
\text{TextF1}@\tau = \frac{2 \cdot \text{TextP}@\tau \cdot \text{TextR}@\tau}{\text{TextP}@\tau + \text{TextR}@\tau}
\end{align}
We report these at $\tau = 0.5$ and $\tau = 0.75$.

\paragraph{Soft metric.} The mean span IoU averages the similarity of each GT group's best-matched prediction,
assigning zero to unmatched GT groups:
\[
\text{mSpanIoU} = \frac{1}{M} \sum_{i=1}^{M}
\begin{cases}
A^{\text{text}}_{i,\pi(i)} & \text{if } G_i \text{ is matched} \\
0 & \text{otherwise}
\end{cases}
\]

\subsection{Image Segmentation}

Image segmentation evaluates the spatial quality of predicted masks, without considering character spans. The
structure is identical to Case 1 with mask IoU replacing span IoU.

\paragraph{Mask IoU.} For each pair $(G_i, \hat{G}_j)$, the standard pixel-level IoU is computed between their
binary segmentation masks:
\[
\text{MaskIoU}(G_i, \hat{G}_j) = \frac{|S_i \cap \hat{S}_j|}{|S_i \cup \hat{S}_j|}
\]

\paragraph{Matching.} An independent $M \times K$ similarity matrix $\mathbf{A}^{\text{mask}}$ is constructed with
$A^{\text{mask}}_{ij} = \text{MaskIoU}(G_i, \hat{G}_j)$, and a separate Hungarian assignment $\sigma$ is computed.
This matching is independent of the text matching in Case 1.

\paragraph{Hard metrics.}
\begin{align}
\text{MaskP}@\tau &= \frac{\mathrm{TP}_\tau}{K}, \quad
\text{MaskR}@\tau = \frac{\mathrm{TP}_\tau}{M}, \quad
\text{MaskF1}@\tau = \frac{2 \cdot \text{MaskP}@\tau \cdot \text{MaskR}@\tau}{\text{MaskP}@\tau + \text{MaskR}@\tau}
\end{align}
where
\[
\mathrm{TP}_\tau = \left|\left\{ i : \text{MaskIoU}(G_i, \hat{G}_{\sigma(i)}) \geq \tau \right\}\right|.
\]
We report these metrics at $\tau = 0.5$ and $\tau = 0.75$.

\paragraph{Soft metric.}
\[
\text{mMaskIoU} = \frac{1}{M} \sum_{i=1}^{M}
\begin{cases}
A^{\text{mask}}_{i,\sigma(i)} & \text{if } G_i \text{ is matched} \\
0 & \text{otherwise}
\end{cases}
\]

\subsection{Joint Correspondence}

Joint evaluation measures whether the model correctly \emph{associates} the right text spans with the right masks.
Unlike Cases 1 and 2, which evaluate each modality independently, this case requires both the text grouping and
the segmentation to be correct for the same prediction.

\paragraph{Joint similarity.} The joint score matrix is computed from the element-wise combination of the text and
mask matrices from Cases 1 and 2. We use the geometric mean:
\[
A^{\text{joint}}_{ij} = \sqrt{A^{\text{text}}_{ij} \cdot A^{\text{mask}}_{ij}}
\]
This penalizes imbalanced predictions: if either the text or mask component is poor, the joint score drops
sharply. For instance, a pair with $\text{spanIoU} = 0.9$ and $\text{MaskIoU} = 0.1$ receives a joint score of
only $0.3$.

\paragraph{Matching.} Hungarian matching is applied to $\mathbf{A}^{\text{joint}}$, producing a third assignment
$\rho$ that is generally different from the text-only ($\pi$) and mask-only ($\sigma$) assignments. This matching
jointly optimizes for both modalities.

\paragraph{Correctness criterion.} A matched pair $(G_i, \hat{G}_{\rho(i)})$ is counted as a true positive only if
\emph{both} individual scores exceed the threshold:
\[
\mathrm{TP}_\tau =
\left|\left\{
i :
\text{spanIoU}(G_i, \hat{G}_{\rho(i)}) \geq \tau
\;\text{and}\;
\text{MaskIoU}(G_i, \hat{G}_{\rho(i)}) \geq \tau
\right\}\right|.
\]
This is strict: a prediction with perfect text but poor segmentation (or vice versa) is counted as incorrect.

\paragraph{Hard metrics.}
\begin{align}
\text{JointP}@\tau &= \frac{\mathrm{TP}_\tau}{K}, \quad
\text{JointR}@\tau = \frac{\mathrm{TP}_\tau}{M}, \quad
\text{JointF1}@\tau = \frac{2 \cdot \text{JointP}@\tau \cdot \text{JointR}@\tau}{\text{JointP}@\tau + \text{JointR}@\tau}
\end{align}
We report these metrics at $\tau = 0.5$ and $\tau = 0.75$.

\paragraph{Soft metric.} The mean joint score averages the geometric-mean similarity over all GT groups:
\[
\text{mJS} = \frac{1}{M} \sum_{i=1}^{M}
\begin{cases}
A^{\text{joint}}_{i,\rho(i)} & \text{if } G_i \text{ is matched} \\
0 & \text{otherwise}
\end{cases}
\]

\section{Dataset curation}

We train on a unified mixture of caption-grounding and instance segmentation data that combines human-annotated benchmarks, converted existing datasets, and automatically constructed supervision (Tab.~\ref{tab:data-stats}). Our caption-grounding data includes GoldG, COCONut-PanCap, and GroundedRef, which provide human-written or synthetic captions paired with object-level masks, while our instance segmentation data includes standard benchmarks such as COCO, ADE20K, COCONut, EntitySeg, converted Roboflow-VL-100 datasets, and two automatically curated sources from PixMo Points and SA-1B. This mixture balances precise mask supervision from established datasets with the dense scenes, long-tail object coverage, and diverse text queries needed to train a grounding model that generalizes across both segmentation-style and caption-grounding settings.

\begin{table*}[!h]
  \centering
  \small
  \caption{Statistics of training data. Image counts are unique images per source and the
  total is their sum, so COCO images shared by COCO, COCONut-PanCap and GroundedRef are counted
  once per source.}
  \resizebox{\linewidth}{!}{
  \begin{tabular}{lcccc}
    \toprule
    {\bf Source dataset}
    & {\bf Num. of images}
    & {\bf Num. of instances}
    & {\bf Annotation Type}
    & {\bf Role} \\
    \midrule

    \multicolumn{5}{c}{\emph{Instance Segmentation Data}} \\
    \midrule
    COCO~\cite{lin2015microsoftcococommonobjects}
      & 117.3K & 860.0K & Human annotated & Standard benchmark \\
    COCONut (Objects365 subset)~\cite{coconut2024cvpr}
      & 242.6K & 3.6M & Human annotated & Standard benchmark \\
    % LVIS~\cite{gupta2019lvis}
      % &  &  & Human annotated & Long-tail categories \\
    EntitySeg~\cite{qi2023highquality}
      & 8.1K & 79.1K & Human annotated & Standard benchmark \\
    ADE20K~\cite{zhou2019semantic}
      & 25.6K & 639.5K & Human annotated & Scene-centric objects \\
    Roboflow-VL-100 (81 datasets)~\cite{robicheaux2025roboflow100}
      & 92.9K & 812.3K & Converted benchmark data & Diverse domains \\
    PixMo Points~\cite{molmov1}
      & 56.8K & 1.7M & Auto-constructed & Dense scenes + diverse text queries\\
    SA-1B~\cite{kirillov2023segment}
      & 71.1K & 249.5K & Auto-constructed & Dense scenes + diverse text queries \\
    \midrule

    \multicolumn{5}{c}{\emph{Caption Grounding Data}} \\
    \midrule
    GoldG~\cite{liu2023grounding}
      & 75.2K & 2.6M & Rewritten + pseudo masks & Human-caption grounding \\
    COCONut-PanCap~\cite{deng2025coconut}
      & 100.1K & 1.3M & Rewritten + GT masks & Human-caption grounding \\
    GroundedRef
      & 60.2K & 230.5K & Synthetic captions + GT masks & Referring and compositional grounding \\
    \midrule

    Total
      & 849.8K & 12.0M & -- & -- \\
    \bottomrule
  \end{tabular}
  }
  \label{tab:data-stats}
\end{table*}

\begin{table*}[t]
\centering
\scriptsize
\setlength{\tabcolsep}{5pt}
\renewcommand{\arraystretch}{1}
\caption{
Statistics of the three image-caption benchmarks we curated or repurposed.
Caption length is measured by whitespace-tokenized words per caption.
Span multiplicity denotes the number of distinct character spans associated with each mask instance.
}
\label{tab:test_set_stats}
\begin{tabular}{lrrrrr}
\toprule
Dataset
& Images
& Captions
& Masks
& Caption length
& Span multiplicity \\
& 
& 
& 
& words/caption
& spans/mask \\
\midrule
Flickr30k
& 779
& 2{,}002
& 6{,}717
& 11.01 $\pm$ 4.36 
& 1.06 $\pm$ 0.26\\

GroundedRef
& 1{,}318
& 2{,}025
& 6{,}193
& 10.34 $\pm$ 4.20 
& 1.04 $\pm$ 0.19 \\

COCONut-PanCap
& 2{,}213
& 2{,}213
& 10{,}821
& 50.84 $\pm$ 16.89
& 1.68 $\pm$ 0.90 \\

\bottomrule
\end{tabular}
\end{table*}

% For the image-category benchmarks, we evaluate on the COCO 2017 validation split (4,952
% images, 36,781 instances, 80 categories), LVIS-minival (4,726 images, 50,536 instances, 802
% categories present in the split), and the EntitySeg validation split (1,190 images, 12,234
% instances, 160 categories present in the split). In each case the full category list of the
% split is concatenated into a single text input.

\subsection{Caption data}

We repurpose three existing datasets into a unified caption-grounding format. GoldG~\cite{liu2023grounding} and COCONut-PanCap~\cite{deng2025coconut} are built from human-written captions and processed with a shared two-stage LLM rewrite pipeline to clean coreference annotations and rewrite captions for consistent grounding. GoldG is derived from Flickr30k and GQA annotations from the Grounding DINO training corpus, with pseudo masks generated from ground-truth boxes using SAM3~\cite{carion2025sam3segmentconcepts}, while COCONut-PanCap uses relabeled COCO panoptic captions with ground-truth masks decoded from panoptic annotations. We also repurpose COCO panoptic segmentation to construct GroundedRef, a synthetic dataset of LLM-generated referring and compositional captions paired with ground-truth masks.

\paragraph{Two-stage caption rewrite.}
GoldG and COCONut-PanCap are both processed with the same LLM-based two-stage rewrite pipeline. In the first stage, \emph{Resolve}, we validate and clean the raw coreference groups. This step removes count-mismatched groups, excludes part mentions from whole-object groups, filters mixed-entity groups, expands groups to include all coreferent mentions, and trims spans to avoid embedded-entity prepositional phrases. In the second stage, \emph{Rewrite}, we minimally rewrite the caption so that only objects represented by annotation groups remain, while preserving actions, spatial relations, and scene context. After rewriting, character spans are reassigned on the rewritten caption. Examples that cannot be rewritten coherently are marked as unrewritable and excluded.

\paragraph{GoldG.}
Our first source is GoldG, the grounding corpus used to train Grounding DINO. We use the Flickr30k and GQA subsets from the Grounding DINO training corpus. Starting from the original GoldG entity annotations, we first group annotations into coreference sets, then apply the shared two-stage rewrite pipeline to clean the groups and rewrite the caption. Because GoldG provides box-level rather than mask-level supervision, we convert each ground-truth box into a pseudo instance mask using SAM3~\cite{carion2025sam3segmentconcepts}, and then serialize the result into our unified format. GoldG therefore provides human-written captions and entity spans, refined through LLM-based cleanup, together with pseudo segmentation masks derived from ground-truth boxes rather than human-drawn masks.

\paragraph{COCONut-PanCap.}
Our second source is COCONut-PanCap, a relabeled version of COCO 2017 validation images with panoptic captions and coreference annotations. Each example contains a caption together with raw coreference groups and panoptic segment metadata. We again apply the shared two-stage rewrite pipeline to clean the groups and rewrite the caption. Unlike GoldG, COCONut-PanCap already provides panoptic segment annotations, so we decode binary masks directly from the RGB-encoded panoptic PNGs and store them as compressed RLE masks. We then convert the processed annotations into the same unified output format.

\paragraph{GroundedRef.}
Our third source, GroundedRef, is a synthetic caption-grounded dataset constructed from COCO 2017 panoptic segmentation using LLM-generated captions. We generate two styles of captions. The first consists of short referring expressions that target a single instance in images containing multiple objects from the same category, requiring the caption to disambiguate the target from similar distractors. The second consists of compositional queries, in which two grounded sub-queries are joined by ``and,'' each describing one object relative to others in the scene. In both cases, the underlying masks come directly from COCO panoptic segmentation, providing ground-truth segment supervision. The text spans are produced by the LLM and then post-processed with automated validation, including word-boundary correction and deduplication. Compared with GoldG and COCONut-PanCap, GroundedRef introduces more controlled and compositional language while retaining precise mask supervision.

\subsection{Instance segmentation data}

We train on a heterogeneous collection of instance segmentation data. This includes standard instance segmentation datasets such as COCO~\cite{lin2015microsoftcococommonobjects}, ADE20K~\cite{zhou2019semantic}, COCONut~\cite{coconut2024cvpr}, and EntitySeg~\cite{qi2023highquality}, as well as 81 instance-oriented datasets from Roboflow-VL-100~\cite{robicheaux2025roboflow100}, which we convert into a unified instance segmentation format by prompting SAM3~\cite{carion2025sam3segmentconcepts} with the bounding boxes. While these datasets provide strong mask supervision, they do not fully cover the diversity of long-tail categories, dense scenes with high object counts, and grounding-style supervision needed for our setting. To complement them, we additionally construct instance segmentation data from PixMo Points~\cite{molmov1} and SA-1B~\cite{kirillov2023segment}, whose point annotations, object discovery signals, and dense visual scenes provide scalable supervision for object-centric grounding.

\paragraph{Pixmo.}
We derive an instance segmentation dataset from the counting subset of PixMo Points, where each image is annotated with object labels and corresponding point locations. Because the original labels include both concrete objects (e.g., car, person) and abstract or non-segmentable concepts (e.g., shadow, reflection), we first filter them with GPT-4o-mini, retaining only labels that refer to physical objects with clear spatial extent. We then normalize verbose labels by removing attribute modifiers, so that, for example, a large red fire truck is mapped to fire truck, yielding canonical category names for downstream processing.
To obtain instance masks, we use SAM3 with two complementary prompting strategies. For each image, we generate candidate masks from the normalized category label (concept-prompted) and from each annotated point (point-prompted). We then assign masks to annotations in two passes. In the first pass, each point is matched to the highest-scoring concept-prompted mask that contains it. Any point left unmatched is assigned in a second pass to the highest-scoring point-prompted mask that contains it. Throughout this process, we discard any candidate mask that covers two or more annotated points, since such masks are likely to merge multiple object instances.

\paragraph{SA-1B images.}
We construct an instance segmentation dataset from a subset of SA-1B (Tars 0--99) by combining vision-language object discovery with SAM3 mask generation under strict point-mask consistency constraints. We first filter SA-1B images to retain those with high object density (200--500 segments), focusing on scenes with many countable instances.
For each selected image, we use GPT-4o-mini in a vision call to enumerate the distinct object categories present in the scene (e.g., dog, car, tree). We then run two independent pipelines on these categories. First, MolmoPoint~\cite{clark2025molmopoint} localizes instances of each category by predicting one point per visible object. Second, SAM3~\cite{carion2025sam3segmentconcepts} generates candidate instance masks from text-only prompts, where category names are normalized by removing quantifier prefixes and converting nouns to singular form (e.g., a group of people becomes person).
We merge the two outputs using a strict validation procedure. For each object category in an image, we require that (1) the number of predicted points matches the number of generated masks, and (2) there is an exact one-to-one correspondence between points and masks, such that each point lies inside exactly one mask and each mask contains exactly one point. Categories that fail either condition are discarded. This agreement between independently produced point localizations and segmentation masks serves as a strong quality filter, retaining only instances that are spatially well-defined and unambiguous.

\section{Baseline implementation details}
\label{app:baselines}

\paragraph{Grounding DINO + SAM.}
Our first baseline is a two-stage pipeline that combines an open-vocabulary detector with a prompted segmentation model. In the first stage, we use Grounding DINO~\cite{liu2023grounding} (\texttt{IDEA-Research/grounding-dino-base}) or its successors MM-Grounding DINO~\cite{zhao2024open} and LLMDet~\cite{fu2025llmdet} to process the input image together with the full caption. Grounding DINO produces a set of detection queries, each associated with a bounding box and a similarity score over caption tokens. We retain detections whose box confidence exceeds $0.25$. For each retained detection, we threshold the token-level similarity logits at $0.20$ and map the activated tokens back to character offsets in the original caption using the tokenizer offset mapping. Adjacent or overlapping character spans are merged, yielding a set of grounded text spans paired with each detected box. This procedure allows the baseline to recover text-region correspondences directly from the detector outputs without using any external language model or post-hoc span generator.
Because Grounding DINO's text encoder has a limited context window, we split long captions exceeding roughly 800 characters (approximately 256 text tokens) into smaller chunks. We split at phrase boundaries using period separators when possible, run detection independently on each chunk, and then remap the predicted character spans back to coordinates in the full caption. The detections from all chunks are merged to form the final set of box--text predictions for the example.

In the second stage, each detected box is passed to SAM3~\cite{carion2025sam3segmentconcepts} for instance segmentation. Unless otherwise noted, we use the bounding box alone as the SAM3 prompt. We also support variants that use the recovered grounded text alone or both box and text jointly as prompts, but the default configuration uses box prompts only because it is the most stable and directly follows the standard detector-to-segmentation pipeline. 

\paragraph{Florence-2.}
We also use Florence-2~\cite{xiao2024florence} (\texttt{florence-community/Florence-2-large-ft}) as a unified model for both localization and segmentation. For caption-based evaluation, we use the \texttt{<CAPTION\_TO\_PHRASE\_GROUNDING>} task, which takes the image and caption as input and predicts a set of bounding boxes paired with phrase labels. Unlike Grounding DINO, Florence-2 does not directly return exact character offsets in the source caption. To obtain text spans comparable to our formulation, we recover character offsets by matching each predicted phrase label back to the original caption. We first perform exact substring matching. If exact matching fails, we apply a normalized token-level matching fallback that ignores minor punctuation and whitespace differences. When multiple detections share the same phrase label, we sort detections by horizontal box center and assign them left-to-right to successive matching occurrences in the caption. To reduce duplicate detections, we apply per-label non-maximum suppression with an IoU threshold of $0.3$.

For segmentation, each detected box is passed to Florence-2's \texttt{<REGION\_TO\_SEGMENTATION>} task, which returns polygon coordinates for the segmented region. We convert these polygons into binary masks and encode them as COCO compressed RLE for evaluation. Because Florence-2 represents locations on a discrete 0--999 coordinate grid, bounding box coordinates are quantized to this grid before prompting the region-to-segmentation head. For category-list style inputs, we instead use Florence-2's \texttt{<OPEN\_VOCABULARY\_DETECTION>} task and query one category at a time. When the query string becomes too long, we split it into smaller category groups; in practice, we chunk queries longer than 2000 characters.

\paragraph{GLaMM.}
GLaMM is a native grounded-segmentation multimodal model that generates a description interleaved with \texttt{<p>phrase</p>[SEG]} markers and emits one segmentation mask per \texttt{[SEG]} token, without a separate detector or mask model. Since GLaMM produces its own grounded description rather than grounding a provided caption, we adapt it to our setting by conditioning it on the target caption—prompting it to segment the objects the caption mentions—and matching each generated phrase back to a character span in that caption (exact match, then token-overlap, then the full caption as a fallback) to recover the text grounding; masks are taken directly from GLaMM. For the image-category setting, generated phrases are matched to the provided category list. We use the GLaMM-FullScope checkpoint.

\paragraph{GPT-5.4 + SAM.}
As a strong proprietary MLLM baseline, we prompt GPT-5.4 to ground each mentioned concept: for the image-caption setting it returns, per concept, the grounded character span(s) in the caption together with a bounding box; for the image-category setting it returns the category name (which we map to its span in the category list) with a bounding box. Each predicted box is then converted to a segmentation mask by SAM3, mirroring the GDINO+SAM pipeline. We run GPT-5.4 with medium reasoning effort.

\paragraph{Qwen3.5-FT.}
For the Qwen3.5 fine-tuning baseline, we use the same Qwen3.5-0.8B backbone and fine-tune it on our correspondence training data with standard autoregressive supervised fine-tuning. Given an image-text pair, the model is trained to generate a serialized list of correspondences, where each entry contains the grounded character span, and a polygon representation of the corresponding image mask. 
Specifically, for each ground-truth instance, we obtain its mask boundary as a polygon and pair it with the character spans of the corresponding phrase in the input text. Masks are first converted into polygon contours. To keep the target representation compact, each polygon component is uniformly subsampled to at most 16 points, and all coordinates are normalized to a fixed integer coordinate range $\{0,\ldots,999\}$. Each target line is formatted as: \texttt{<x1 y1 x2 y2 ... x16 y16 | x1 y1 x2 y2 ... x16 y16 ; s e>} where polygon components belonging to the same instance are separated by \texttt{<|>}, the semicolon separates visual geometry from textual grounding, and \texttt{<s e>} denotes the character span of the grounded mention. Multiple instances are serialized as separate lines following the ground-truth instance order. 
We use a maximum training sequence length of 8,192 tokens and a 16K-token generation budget during evaluation to accommodate long multi-instance outputs.

This baseline tests whether a pretrained VLM can learn bidirectional correspondence purely through language-generation supervision, without our bridge-token correspondence hypotheses or dedicated text and image segmentation heads. Its strong text scores show that autoregressive fine-tuning can recover grounded mentions reasonably well, but its weaker mask and joint metrics indicate that generating structured masks as text is less effective for precise instance-level correspondence prediction.

\section{Effects of Annotation Standardization and Model Architecture}
\label{sec:annotation_architecture_analysis}

We further analyze the effects of annotation standardization and model architecture by training \model and MM-GDINO with either the original source annotations or our standardized annotations. Both models are fine-tuned for 15K steps. Within each model comparison, all other training and evaluation settings are kept unchanged. \model uses its native dynamic-resolution input.

\begin{table}[t]
\centering
\scriptsize
\setlength{\tabcolsep}{5pt}
\renewcommand{\arraystretch}{1.05}
\caption{Effects of annotation standardization and model architecture. Image-caption JointF1 is averaged over the three caption benchmarks, while image-category JointF1 is averaged over COCO and LVIS-minival.}
\label{tab:data_arch_contributions}
\begin{tabular}{llcc}
\toprule
\textbf{Training annotations} &
\textbf{Method} &
\textbf{Image-caption} &
\textbf{Image-category} \\
& & \textbf{JointF1} & \textbf{JointF1} \\
\midrule
Original source
& MM-GDINO-FT + SAM & 57.0 & 33.3 \\
Original source
& \textbf{\model} & \textbf{60.8} & \textbf{44.9} \\
\midrule
Standardized
& MM-GDINO-FT + SAM & 66.4 & 33.6 \\
Standardized
& \textbf{\model} & \textbf{70.6} & \textbf{45.2} \\
\bottomrule
\end{tabular}
\end{table}

\paragraph{Effect of annotation standardization.}
Our standardized image-caption annotations recover omitted grounded entities, consolidate coreferential mentions, and align each grounded text span with its corresponding instance mask. This improves image-caption JointF1 by 9.4 points for MM-GDINO-FT and 9.8 points for \model.

\paragraph{Effect of model architecture.}
With the original annotations, \model outperforms MM-GDINO-FT by 3.8 points on image-caption benchmarks and 11.6 points on image-category benchmarks. With the standardized annotations, the corresponding gains are 4.2 and 11.6 points. The consistent improvements under both annotation settings show that the architectural advantage is complementary to the gains from annotation standardization.

\section{Visualizations}

\begin{figure}[t!]
    \centering
\includegraphics[width=\textwidth]{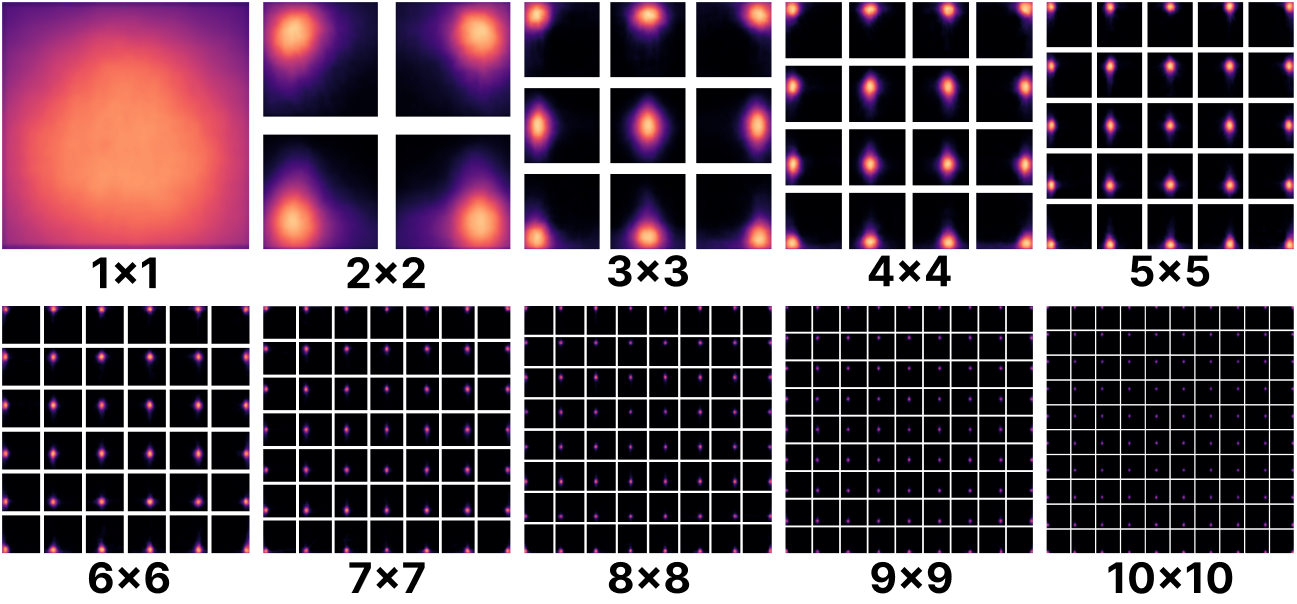}
\caption{
    \textbf{Spatial specialization of bridge tokens across token groups.}
    We visualize the spatial activation patterns of bridge tokens from different scale groups under our multi-scale bridge-token design.
    Each group corresponds to a spatial partition of the image, ranging from a global $1\times1$ group to increasingly fine grids.
    Bridge tokens in coarse groups respond to broad image regions, while tokens in finer groups activate on increasingly localized areas aligned with their assigned spatial cells.
    This shows that bridge tokens learn structured spatial priors and progressively cover the image at multiple resolutions, supporting correspondence prediction for objects of different sizes and locations.
    }
    \label{fig:bridge-act}
\end{figure}

\textbf{Bridge token spatial specialization.}
We next visualize the activation patterns of bridge tokens from different token groups to understand the effect of our multi-scale spatial assignment.
As shown in Fig.~\ref{fig:bridge-act}, bridge tokens in the coarse groups (e.g., $1\times1$ and $2\times2$) produce broad activations that cover large regions of the image, whereas tokens in finer groups become progressively more localized and align with smaller spatial cells.
This behavior is consistent with the intended design of our bridge tokens: different groups encode different spatial scales, and individual tokens specialize to particular regions within those scales.
The visualization suggests that the model learns a structured set of correspondence queries that jointly cover the image from global to local resolution, which helps it handle objects with diverse sizes and positions.

\begin{figure*}[t]
    \centering
    \includegraphics[width=\textwidth]{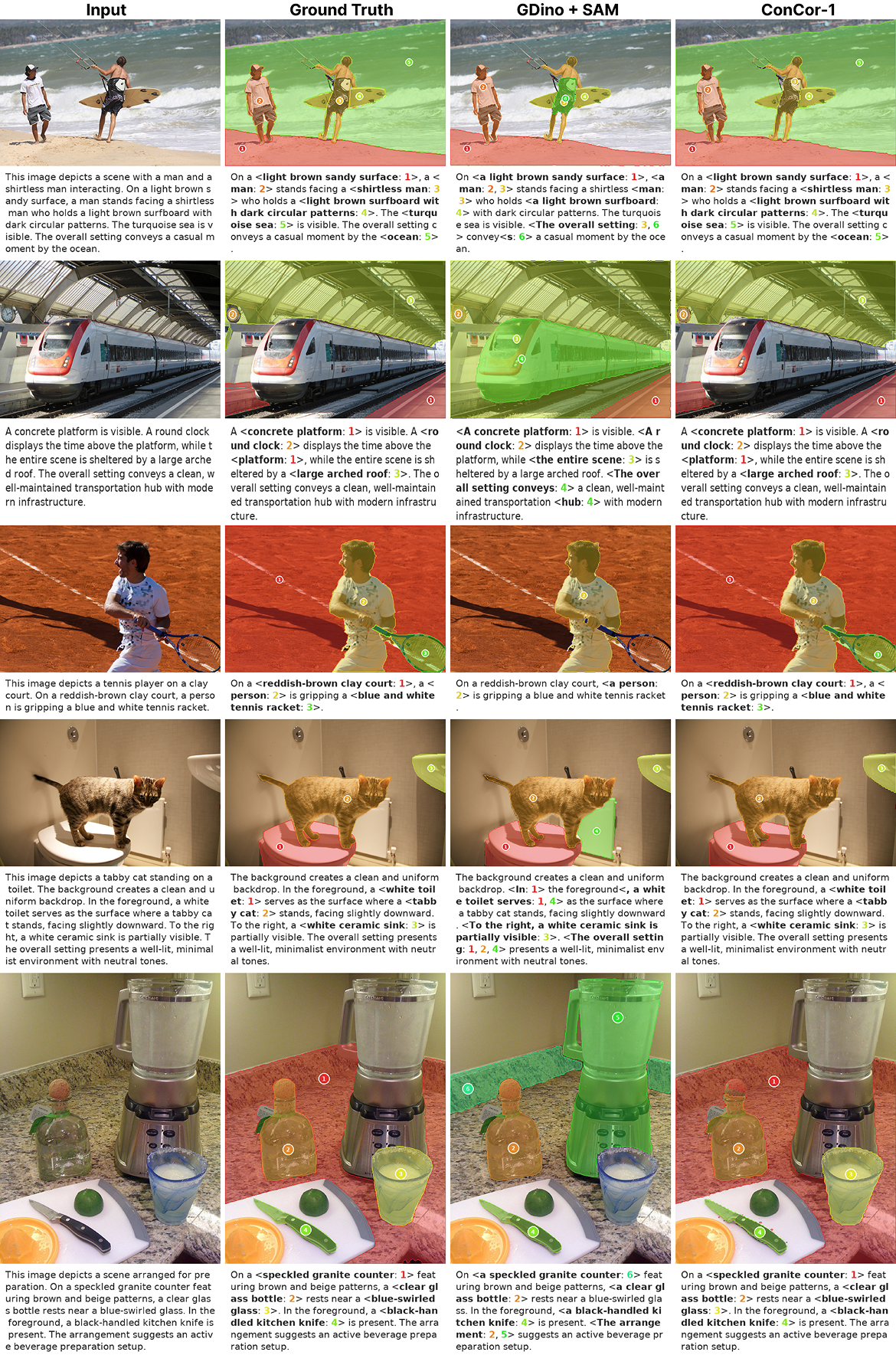}
    \caption{Qualitative visualization examples.}
    \label{fig:examples-compare-app}
\end{figure*}

\textbf{Additional qualitative comparisons.}
Fig.~\ref{fig:examples-compare-app} provides additional qualitative comparisons on COCONut-PanCap. Across diverse scenes with multiple objects and complex captions, GDINO+SAM often detects visually plausible regions but fails to associate them with the correct caption mentions. This is especially common when the image contains multiple instances of the same category or when the referred object is described through contextual relations. In contrast, \model more consistently recovers both the correct text spans and the corresponding image masks, producing more complete and instance-specific text--image alignments.

\begin{figure*}[t]
    \centering
    \includegraphics[
        width=\textwidth,
    ]{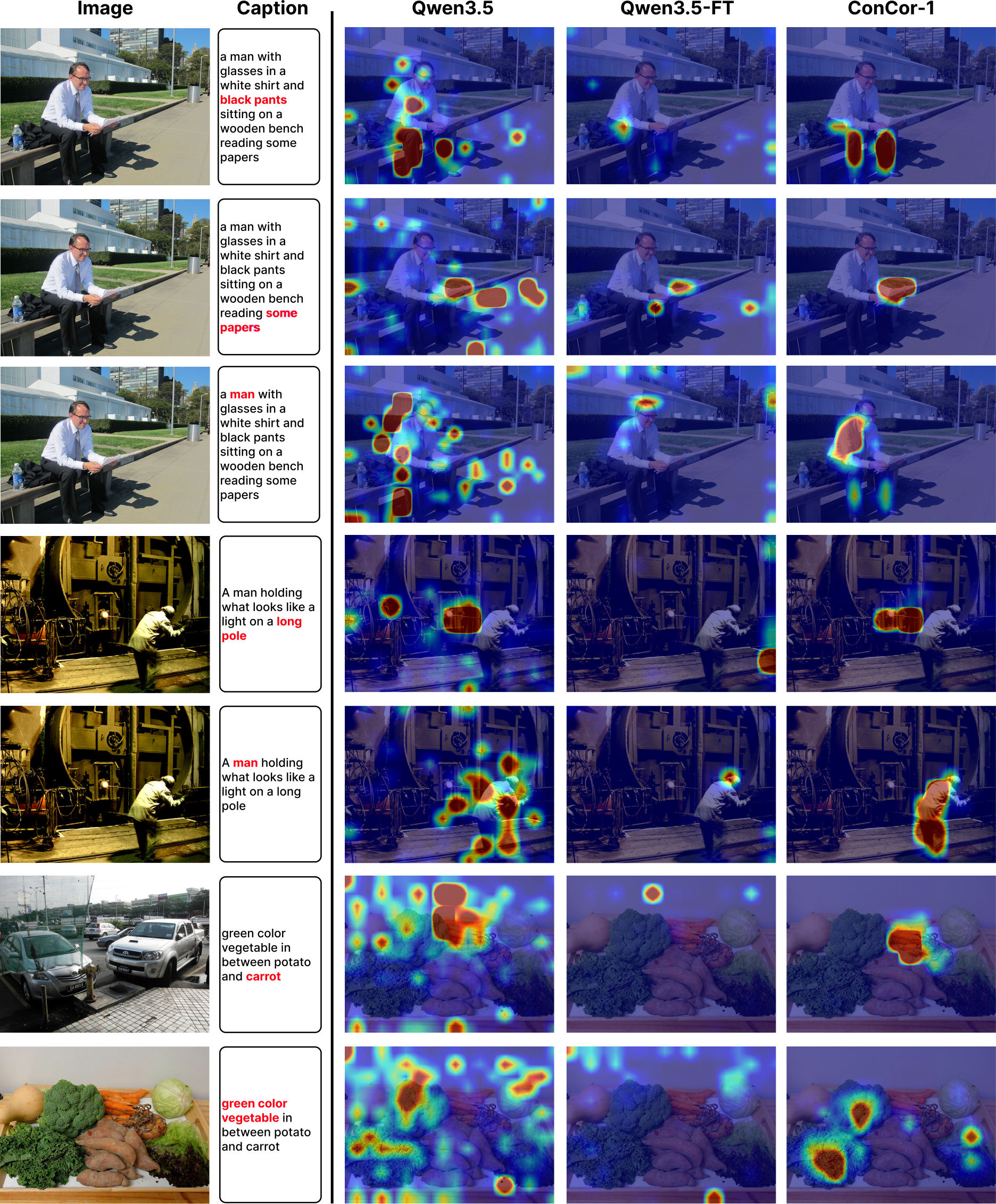}
    \caption{Qualitative visualization examples.}
    \label{fig:t2v-attention-app}
\end{figure*}

\textbf{Additional text-to-image attention visualizations.}
Fig.~\ref{fig:t2v-attention-app} shows additional layer-20 text-to-image attention maps for the original Qwen3.5, Qwen3.5 fine-tuned on our dataset, and \model. The original model often attends broadly to background regions or nearby distractor objects, while fine-tuning on our data improves localization but still produces noisy attention in many cases. \model yields more compact and semantically meaningful attention maps, where object words attend to the corresponding visual entities. These examples further support that the bridge-token correspondence objective improves not only the final grounding outputs, but also the internal cross-modal alignment of the VLM.

\begin{figure*}[t]
    \centering
    \includegraphics[
        width=\textwidth,
    ]{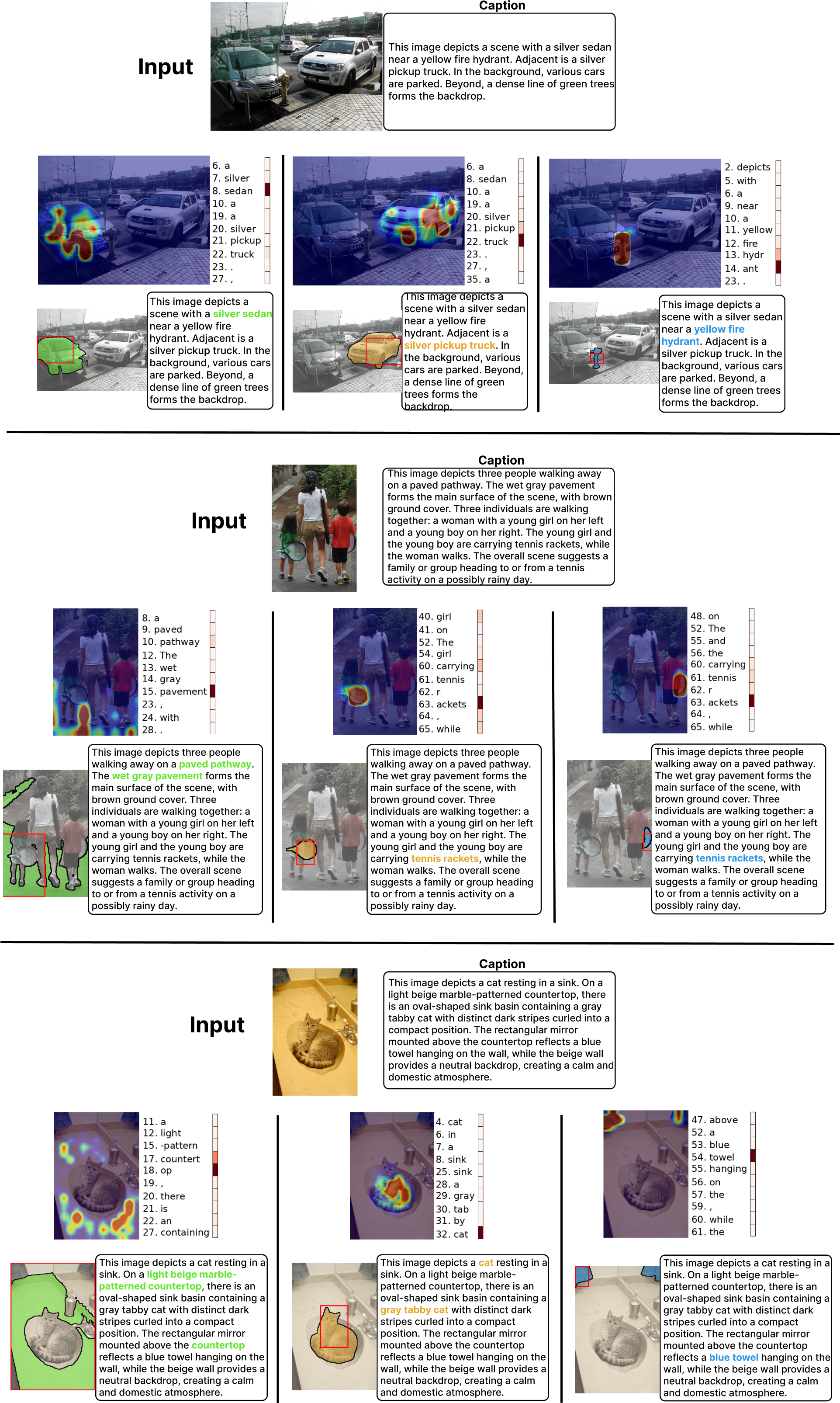}
    \caption{Qualitative visualization examples.}
    \label{fig:bridge-att-app}
\end{figure*}

\textbf{Additional bridge-token attention visualizations.}
Fig.~\ref{fig:bridge-att-app} presents additional examples of bridge-to-image and bridge-to-text attention, together with the spatial prior and final prediction for each selected bridge token. The visualizations show that different bridge tokens specialize to different object-level hypotheses: each token tends to focus on a localized image region and the corresponding text mentions. The final predictions demonstrate that these bridge-specific attention patterns are effectively transformed into coherent text--mask and image--mask correspondences. This provides further evidence that bridge tokens serve as structured intermediate variables for binding language spans with visual instances.

\clearpage
\clearpage

\clearpage

\clearpage

%%%%%%%%%%%%%%%%%%%%%%%%%%%%%%%%%%%%%%%%%%%%%%%%%%%%%%%%%%%%

% \clearpage
% \input{checklist.tex}

\end{document}